\documentclass[twocolumn]{article}  % Use the two column option

\usepackage[left=0.75in, right=0.75in, top=0.75in, bottom=0.75in]{geometry}

\usepackage{graphicx}  % For including images
\usepackage{amsmath}   % For math formatting
\usepackage{hyperref}  % For hyperlinks
\usepackage{orcidlink}
\usepackage{mathptmx}
\newcommand{\task}[1]{T#1}%{T\textsubscript{#1}}
\usepackage{comment}

\title{\textbf{XMTC: Explainable Early Classification of Multivariate Time Series in Reach-to-Grasp Hand Kinematics}}
\author{
    Reyhaneh~Sabbagh Gol$^{1}$, Dimitar~Valkov$^{2}$, and Lars~Linsen$^{1}$ \\
    $^1$University of Münster, Germany \\
    $^2$Saarland University, Saarbrücken, Germany
}
\date{}  % Remove date from title

\begin{document}
\maketitle

\begin{abstract}
Hand kinematics can be measured in Human-Computer Interaction (HCI) with the intention to predict the user's intention in a reach-to-grasp action. Using multiple hand sensors, multivariate time series data are being captured. Given a number of possible actions on a number of objects, the goal is to classify the multivariate time series data, where the class shall be predicted as early as possible. 
Many machine-learning methods have been developed for such classification tasks, where different approaches produce favorable solutions on different data sets. We, therefore, employ an ensemble approach that includes and weights different approaches. 
To provide a trustworthy classification production, we present the XMTC tool that incorporates coordinated multiple-view visualizations to analyze the predictions. 
Temporal accuracy plots, confusion matrix heatmaps, temporal confidence heatmaps, and partial dependence plots allow for the identification of the best trade-off between early prediction and prediction quality, the detection and analysis of challenging classification conditions, and the investigation of the prediction evolution in an overview and detail manner.
We employ XMTC to real-world HCI data in multiple scenarios and show that good classification predictions can be achieved early on with our classifier as well as which conditions are easy to distinguish, which multivariate time series measurements impose challenges, and which features have most impact. 

\end{abstract}

\section{Introduction}
%time series
Time series data, i.e., data that are collected over time, are gathered in a wide range of fields such as healthcare~\cite{clemente2020helena}, finance~\cite{idrees2019prediction}, economics~\cite{marcellino2007comparison}, energy~\cite{zhen2021photovoltaic}, climate~\cite{karevan2020transductive}, and Human Computer Interaction (HCI)~\cite{ruiz2024children,blalock2016extract, subasi2025feature, shin2024methodological, zerrouki2024deep}. 
% time series tasks e.g. classification
A wide range of deep learning and ensemble models have been employed for processing time series data in various tasks such as classification, forecasting, regression, and generation of time series~\cite{middlehurst2021hive, ismail2020inceptiontime, tan2022multirocket, mohammadi2024deep, wen2023transformers, ahmed2023transformers,song2018attend,cabello2024fast}.
%multivariate time series
Much research has been conducted focusing on univariate time series classification (TSC), where each instance consists of a single time series. However, many real-world problems, such as human activity recognition, hand tracking, gesture recognition, medical diagnosis using electrocardiogram (ECG), electroencephalogram (EEG), or magnetoencephalography (MEG), and systems monitoring, are mostly multivariate. Despite their importance in everyday life, multivariate time series classification (MTSC) has received significantly less attention than univariate time series~\cite{ruiz2021great}. Moreover, most univariate and multivariate time series classification models are designed for equal-length, synchronized time series, which often do not reflect real-world scenarios. 

%our prediction model 
In this paper, we present an approach for the early prediction of classification results from non-synchronized multivariate time-series data with a particular emphasis on the explainability of the classification prediction. 
Machine-learning models that are designed for processing the time series data typically prioritize accuracy over transparency and interpretability of black-box models. Lack of interpretability and explainability causes many critical challenges in important decision-making contexts and raises concerns about the reliability of the black-box systems among analysts and decision-makers. The research field of eXplainable AI (XAI) explains the models' behavior to give insights and interpretation of the models' decisions to users~\cite{kamath2021explainable, dwivedi2023explainable, abusitta2024survey, zhao2023interpretation, adadi2018peeking, theissler2022explainable, zhao2023interpretation}. 

As many classification methods for time series exist, where the best choice is a priori not known, we decided to employ a state-of-the-art ensemble classifier that comprises of several classifiers of different types and combines them in a weighted fashion~\cite{middlehurst2021hive}. As such an approach that learns the weights is expensive, we identify the dominant classifier and continue with that approach without loss of accuracy. The identified classifier is itself an ensemble classifier called \textit{Diverse Representation Canonical Interval Forest (DrCIF)}~\cite{middlehurst2021hive} specifically designed for time series data. DrCIF is an interval-based classifier that finds discriminatory features over different intervals. Precisely, DrCIF constructs an ensemble of time series trees by extracting diverse features from randomly selected intervals across three representations—base series, first-order differences, and periodograms—and combining them with a candidate pool of 29 statistical features, ensuring diversity through randomization of intervals, dimensions, and feature subsets for each tree~\cite{middlehurst2021hive}.

%XAI
% Partial Dependence Plot
Our approach for eXplainable Multivariate Time series Classification (XMTC) embeds the classification prediction into a visual analysis tool for the interpretation of classification results. XMTC allows for (1) the detection of a good trade-off between early prediction and accuracy via temporal accuracy plots, (2) the investigation of the classification output at each point in time via confusion matrices, (3) the detailed investigation of the temporal evolution of the classification prediction for individual multi-variate time series, and (4) the analysis of the impact of input features on the classification output via a Partial Dependence Plot (PDP)~\cite{kamath2021explainable}.
%offers insights into the overall behavior of the model by illustrating the relationships between input features and the output, offering insights into the global behavior of the model by illustrating the relationships between input features and the output. PDP depicts the partial dependence function, which evaluates a feature's effect by marginalizing over the other features~\cite{kamath2021explainable}.

%HCI, R2G, object predition in R2G
We have developed our approach within the context of predicting the users' intention in a reach-to-grasp hand motion.
The rise of machine learning approaches, virtual reality, and augmented reality has renewed the interest in hand- and gesture-based interfaces, enabling intuitive object manipulation. As in real life, hands are used to explore the surrounding environment, grasp objects, and perceive the shape and weight of objects. Thus, research fields like hand tracking~\cite{taylor2016efficient} and gesture recognition~\cite{cheng2015survey} by using hand- or body-attached sensors~\cite{chen2016finexus} have been getting more attention. 
%Therefore, the research area focusing on the hand's ability to convey environmental and perceptual information, such as an object's shape, size, and orientation through finger interactions, receives less focus. 
Reach-to-grasp (R2G) is the gradual modeling of temporal and spatial information of the user's hand to grasp an object. Although hand tracking concerning Fitts' law has been extensively studied, prediction of the intended object in R2G according to temporal behavior of fingers (prehensile kinematics) has not been given much focus~\cite{valkov2023reach}. 
% our work
We address this challenge by applying XMTC to R2G data, which have been acquired in a controlled data acquisition study to gather high-precision data on prehensile hand movements.

Our main contribution can be summarized as follows:
\begin{itemize}
    \item The design of the XMTC tool for interactive visualization, exploration, and evaluation of classification prediction models for non-synchronized multivariate time series.
    \item Adoption of an interval-based classifier model to gradually predict the classification results with increasing length of the time series.
    \item Interactive visual estimation of a trade-off point between early prediction and classification accuracy.
    \item Interactive visual exploration of the temporal evolution of the classifier models with increasing length.
    \item Detailed investigation of the classification prediction evolution of individual multi-variate time series.
    \item Detailed analysis of the impact of input features on the individual model's classification output.
    \item Application of the methods to R2G data for detecting user intentions, which 
    % \item Provide insights to the user by plotting confusion matrices over time during the R2G process, highlighting similar and confusing classes that the model misclassifies.
    % \item Exploring predictive models to identify the most suitable one that accurately classifies objects at the earliest possible stage.
    % \item Gradually explore model predictions for each individual test time series during the R2G process.
    % \item Observing the global behavior of the model and Finding distinguishable input features using PDP plots for input features.
    % \item Evaluate the model's generalizability for unknown users using the leave-one-out test.
%These findings could guide the design of future interactive environments by enhancing the discriminability of virtual objects. Moreover, they 
offers valuable insights for developing interfaces that use everyday objects as haptic proxies~\cite{valkov2023reach} and open the door to utilizing predictive algorithms to account for latency and anticipate future actions.
\end{itemize}

\section{Background and Related Work}
\label{sec:related}
\textbf{Hand prehension analysis.}
Human-Computer Interaction (HCI) is the research field that focuses on making technological interaction easy for the user~\cite{nazar2021systematic}. It has been getting a lot of focus by emerging new technologies in the modern world. Hands play an important role in user interactions, and hand prehension is one of the most important domains in HCI, psychology, and neuroscience~\cite{valkov2023reach}. Some studies that focused on hand preshaping in R2G tasks proved that grasping information depends on the object's size, shape, and intended action~\cite{betti2018reach, egmose2018shaping}. Some papers studied the relationship between hand transport and grasp formation. They found that, while hand transport and grasp formation are largely independent, pre-shaping is typically completed before the hand reaches the object~\cite{chieffi1993coordination, ansuini2015predicting, santello1998gradual, molina2002neural, daiber2012towards}. Most of these work rescale, trim, or resample the motion sequences which is not practical in real-time. Valkov et al.~\cite{valkov2023reach} attempted to classify the intended object using reach-to-grasp hand kinematics. However, their network failed to reliably discriminate the synthetic object, achieving performance close to chance level. In this paper, we predict the intended object in the reach-to-grasp task, before grasping the object and without synchronizing or trimming motion sequences.

\noindent
\textbf{Machine learning for MTSC.}
While a wide range of Machine-learning approaches classify univariate time series, multivariate time series have received less focus. We refer to recent surveys in the field~\cite{middlehurst2024bake, wen2023transformers, ahmed2023transformers}. Middlehurst et al.~\cite{middlehurst2024bake} categorize the different approaches into methods based on distance, feature, interval, shapelet, dictionary, convolution, and deep learning as well as hybrid ones. As the best approach may vary from application to application, meta-ensemble approaches have been introduced that combine several approaches and weigh them, where the weights are learned. \textit{Hierarchical Vote Collective of Transformation-based Ensembles v2 (HIVE-COTE2)} is a state-of-the-art meta ensemble for time series classification consisting of four classifiers of diverse categories identified by Middlehurst et al. It includes phase-independent shapelets, bag-of-words-based dictionaries, phase-dependent intervals, and convolution-based classifiers. Each classifier is trained independently to estimate the probability of class membership for unseen data. The control unit then combines these probabilities, weighting them based on an estimate of the module's performance on the training data~\cite {middlehurst2021hive}. We follow this approach by employing HIVE-COTE2 to our data. However, due to the large computation times, we investigated the learned weights and then focused on the most influential classifier only, which was possible without a loss of accuracy.

\noindent
\textbf{Explainable AI.}
Advances in machine-learning algorithms have significantly enhanced predictive power and accuracy. However, this progress has come at the cost of increasing complexity, presenting a challenging trade-off between performance and transparency. The field of Explainable AI (XAI) addresses the explainability and interpretability of machine learning and deep learning algorithms~\cite{kamath2021explainable}.
For recent surveys in XAI, see, e.g.,~\cite{dwivedi2023explainable, abusitta2024survey, adadi2018peeking} and, for the recent surveys in XAI for time series, see, e.g.,~\cite{theissler2022explainable, zhao2023interpretation}.

Kamath et al.~\cite{kamath2021explainable} categorize XAI methods based on their scope, stage, and model type. 
XAI methods can be categorized as global or local in \textit{scope}. Global methods are valuable for interpreting the overall, macro-level behavior of models, while local methods are useful for understanding their behavior at a micro-level, focusing on individual predictions.
XAI methods can be divided into three categories based on \textit{stage}: pre-model, intrinsic, or post-hoc methods. This is based on whether they are applied before, during, or after a model makes its prediction. 
When investigating the \textit{model type}, methods can be model-specific or model-agnostic. Model-specific methods utilize unique characteristics or the architecture of a particular model to achieve explainability. While many pre-model and post-hoc explainability techniques are model-agnostic and can be applied to a broad range of models, some are tailored to specific types of models (e.g., convolutional neural networks) and are limited to their respective architectures.
Model-agnostic methods have the advantage that they can be implemented and used for any model and can compare different models together. The advantage of post-hoc methods is that they can be applied to any trained black-box model, i.e., they can work for a wide variety of model algorithms, without the need for understanding their internal structure.
We conclude that a global, post-hoc, and model-agnostic method is most desirable for our purposes, as it gives the user some global insights and is comparable and applicable to any other trained models. 

Some well-established post-hoc methods are Shapely values, Local Interpretable Model-Agnostic Explanations (LIME), Partial Dependence Plots (PDP) , confusion matrices, and global surrogates.
Shapely values are based on the game theory and compute the (weighted) average of the marginal contributions of each player (i.e., feature value) across all possible coalitions~\cite{abusitta2024survey}. 
Local Interpretable Model-Agnostic Explanations (LIME) is a model-agnostic technique for generating interpretable explanations of individual predictions in the vicinity of the prediction. This approach operates by randomly perturbing instances and fitting a local surrogate model to these perturbations~\cite{kamath2021explainable}.
Partial Dependence Plots (PDP), as introduced by Friedman~\cite{friedman2001greedy}, quantify the marginal impact of one or more features on a machine-learning model's predicted outcome. It is among the most widely used model-agnostic techniques~\cite{angelini2023visual}.
Confusion matrices are commonly used to visualize classification results on the test dataset. They provide quantitative metrics to assess the model's generalization performance and serve as a diagnostic tool to analyze its behavior on individual classes~\cite{kamath2021explainable}.
The global surrogate model aims to explain a complex model by approximating the predictions with a simpler, more interpretable model. The simpler model provides insights into the global behavior of the complex model. Global surrogate models typically lack the ability to provide precise predictions for individual data instances. Even when a surrogate model accurately mirrors the predictions of a black-box model, it does not guarantee that these predictions correspond to real-world outcomes~\cite{kamath2021explainable}.

In this paper, we require a global, post-hoc, and model-agnostic approach that fulfills a number of tasks. For detecting the trade-off point between prediction earliness and accuracy, we develop a temporal plot, which is combined with confusion matrix visualizations to investigate how the models evolve with increasing time-series lengths. Confusion matrices provide a good global overview of the classification output. For the investigation of the influence of individual features on the classification output, Partial Dependency Plots have been developed, as explained above, and we follow this approach. For the detailed investigation of the classification evolution for individual multi-variate time series, we adopt the heatmap visualization proposed by Nourani et al.~\cite{nourani2022detoxer}.

\section{Hand Kinematics Time Series}
\label{sec:data_collection}
Our goal is to predict the users' intention when interacting with objects already during the reach-to-grasp phase. 
Therefore, we collected hand and finger motion tracking data % for the \textit{reach-to-grasp(R2G)} phase and \textit{perform-action} phase 
for 29 adults. 
Three different experiments were conducted as detailed below.
In all three experiments, users should reach the objects and then perform an action. Each participant tries the experiment in two trials. Objects are located in one of two positions, 30 cm to the right or left side of the participant. 

\noindent
\textbf{Experiment 1}: The participants had to reach out and grasp one of four \textit{real objects} and perform an action with it. The combination of objects and actions is detailed in Table~\ref{table:objects_actions_task1}.

\begin{table}[ht]
\centering
\begin{tabular}{|c|c|c|c|c|}
\hline
\textbf{Action Name} & \textbf{Bottle} & \textbf{Cup} & \textbf{Knife} & \textbf{Pen} \\ \hline
Non-use             & Move            & Move         & Move           & Move         \\ \hline
Primary use         & Pour            & Drink        & Cut            & Write        \\ \hline
Secondary use       & Drink           & Pour         & Poke           & Poke         \\ \hline
\end{tabular}
\caption{The combination of object and actions for Experiment 1.}
\label{table:objects_actions_task1}
\end{table}

\noindent
\textbf{Experiment 2}: The participant had to reach out and \textit{grasp}, \textit{touch}, or \textit{push} one of four \textit{synthetic objects}. The synthetic objects were a \textit{bar} (a high prismatic object), a \textit{box} (a medium-sized cube), a \textit{dice} (a small cube), and a \textit{plank} (a 2cm thin plank).

\noindent
\textbf{Experiment 3}: The participant had to reach out and \textit{grasp} one of the \textit{synthetic objects} mentioned above and perform an action with it. The actions were
 \textit{center} (place the object precisely at the predefined position),  \textit{somewhere} (place the object somewhere on the table),  \textit{hold} (hold the object for 2 seconds above the table), and \textit{throw} (throw the object in a trash bin beneath the table).

For the experiments, we used a right-handed coordinate system with the x-axis pointing upwards, the y-axis pointing forward, and the z-axis pointing from right to left. The origin is at the \textit{hand starting position}. All positions were implemented as touch-sensitive surfaces. These were 3D printed with conductive carbon-based filament (ProtoPasta CDP1175) and connected to a touch driver MPR121 and an Arduino Nano33 IoT board that captured the touch events at 250 fps.
We utilized the Polhemus Viper16 electromagnetic tracking system, equipped with 12 tethered microsensors, to monitor participants’ hands and fingers with submillimeter accuracy at a sampling rate of 960 fps. Five sensors were affixed to the fingernails, while another five were positioned in the middle of the proximal phalanges of the fingers. Additionally, sensors were attached to the metacarpal of the thumb and the metacarpal of the middle finger. %The data acquisition application was developed in C/C++ using the Qt framework (v6.3) and executed on a Windows 11 laptop powered by a Core i7-11800H processor, 32 GB of RAM, and an Nvidia RTX 3080 GPU. 
Synthetic objects were fabricated using 3D-printed conductive filament, while real objects were augmented with copper tape to facilitate touch transfer to their surfaces. Despite its simplicity, the setup enables high-fidelity motion tracking, reliably capturing key events such as when participants lift their hand from the start position or interact with an object.

Instead of using the sensor data as acquired in a global coordinate system, we compute a feature vector consisting of the main features that were considered to be more meaningful for the R2G gesture. We, therefore, extract the aperture vector between the thumb and each of the other fingers and refer to the features as \textit{Thumb-Index Aperture (tia)}, \textit{Thumb-Middle Aperture (tma)}, \textit{Thumb-Ring Aperture (tra)}, and \textit{Thumb-Little Aperture (tla)}. For each aperture vector, the $x$, $y$, and $z$ coordinates are used as input, resulting in a 12-dimensional time series. The time series may have different lengths and are not synchronized, as synchronization would not be possible in the real-world scenario, where the user's intention is supposed to be predicted during interaction, i.e., when dealing with streaming data. The only preprocessing step that we conducted was a normalization of the feature values to a range $[0,1]$, which is a global scaling that can be applied to streaming data.

\section{Classification}
\label{sec:model}
Given the multivariate time series data representing the feature vectors from the three experiments explained in Section~\ref{sec:data_collection}, we now explain how we define a model for the classification task. 

Since the motivation of this paper is to predict the intended object as early as possible, we employed a moving window approach with an initial window size of 10 time steps, which then is incrementally increased in steps of 10. Thus, the initial window covers time step intervals [0,10], followed by [0,20], [0,30], and so on. 
Thus, we are creating time series datasets of increasing length, where the $i$-th time series dataset captures the time steps from 0 to $i\cdot 10$.
Then, each time series dataset is fed to a classification model to train it.
As the collected time series may finish early, while classification models often require time series of the same length, we stretch too short time series accordingly for the training phase. We then train a classification model for each time series dataset (i.e., for time steps 0 to $i\cdot 10$).

% HIVE-COTE 1 and 2
As discussed in Section~\ref{sec:related}, many different approaches to the classification of time-series data exist. Thus, we follow the idea of training a meta-ensemble classifier model that incorporates several classifiers of different types.
HIVE-COTE (Hierarchical Vote Collective of Transformation-based Ensembles) is an advanced ensemble framework for time series classification that combines diverse types of classifiers to achieve state-of-the-art performance. Its initial version (HIVE-COTE 1.0) integrated classifiers from multiple time series representation categories, including shapelets, elastic distances, and interval-based methods, using a weighted voting scheme. HIVE-COTE 2.0 improved upon this idea by introducing the Collective of Aggregated Weighted Probabilities Ensemble (CAWPE) for weighting classifiers based on their estimated accuracy on unseen data, as well as incorporating new classifiers.

\begin{figure}[htb!]
    \centering
    \includegraphics[width=1\linewidth]{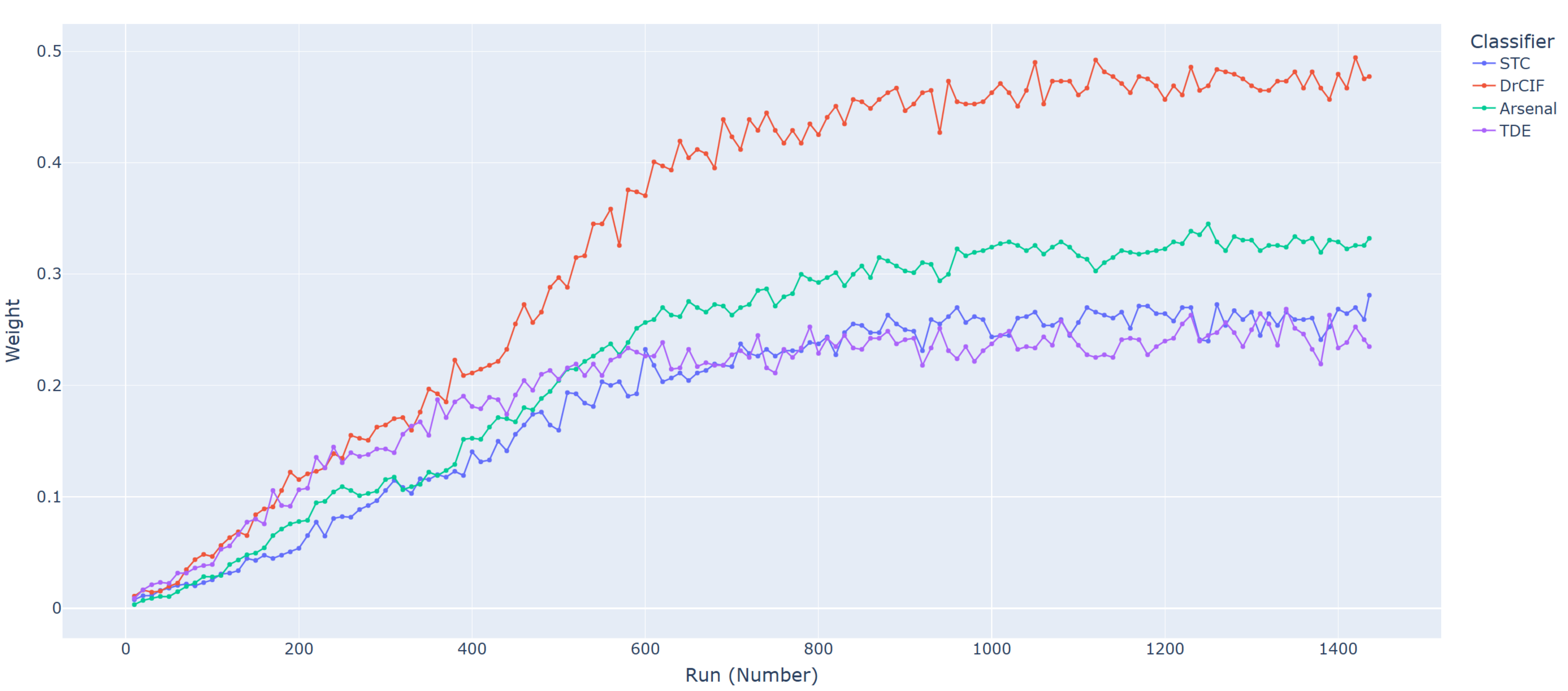}
    \caption{Weights (y-axis) of the four classifiers in HIVE-COTE2 shown for all trained models for increasing time series lengths (x-axis). The plot demonstrates that DrCIF (red) receives higher weights across all models over time when compared to the other three classifiers.}
    \label{fig:hivecote_weights}
\end{figure}

% HIVE-COTE2 is expensive
HIVE-COTE2 is computationally intensive and memory-demanding, making it impractical and inefficient.
In our object prediction approach, we trained HIVE-COTE2 and found that the interval-based classifier, Diverse Representation Canonical Interval Forest (DrCIF), received the highest weight among the four classifiers~\ref{fig:hivecote_weights}. Moreover, DrCIF requires significantly less training time and memory while achieving the same accuracy. Therefore, we decided to use DrCIF for multivariate time series prediction.

Diverse Representation Canonical Interval Forest (DrCIF)~\cite{middlehurst2021hive} is a state-of-the-art ensemble classifier specifically designed for time series data. It builds decision trees by considering intervals of the time series, rather than individual data points. It works by partitioning the data into intervals and learning decision trees based on the distribution of these intervals across the time series, which enhances its ability to capture temporal dependencies. A decision tree, in essence, is a predictive model that splits data into subsets based on feature values to make decisions, represented as a tree structure, where each non-leaf node corresponds to a feature decision and each leaf node to a class label.

More precisely, DrCIF constructs an ensemble of time-series trees by extracting diverse features from randomly selected intervals across three time-series representations and combining them with a candidate pool of 29 statistical features, ensuring diversity through randomization of intervals, dimensions, and feature subsets for each tree~\cite{middlehurst2021hive}.
To do so, the features for each tree are derived from multiple intervals randomly sampled from three distinct representations of the time series: the original series, the first-order difference series, and the periodograms of the full series. From these intervals, seven basic summary statistics (mean, standard deviation, slope, median, interquartile range, minimum, and maximum) are extracted. DrCIF enhances this set by including the catch22~\cite{lubba2019catch22} feature set, expanding the candidate pool to 29 features. For each tree, a subset of size $a$ of these 29 features is randomly selected. For the three representations, $k$  phase-dependent intervals with randomly determined positions and lengths are extracted, and the selected features are computed for each interval. These features are then combined into a feature vector of length $3 \cdot a \cdot k$ for each series, which is used to train the tree. To ensure diversity among base classifiers, each is provided with a unique combination of intervals and features. The number of intervals $k$  is typically chosen based on the representation series length, which varies across representations. For example, the periodogram is half the length of the original series, and the difference series has one value less, leading to potential variation in the number of intervals selected. In the case of multivariate data, the dimension used for each interval is randomly chosen by DrCIF.

In this paper, the DrCIF model was applied with its default settings and parameters, which generates 200 ensemble members. For each ensemble member, 10 attributes are randomly selected from the pool of 29 attributes. Each ensemble member is a decision tree, and the input features for each tree are derived from multiple intervals that are randomly sampled from three distinct representations of the time series. The final decisions from all individual decision trees are aggregated using majority voting.

\section{XMTC}
In the following, we introduce our explainable multi-variate time series classification prediction tool XMTC. We first describe the task analysis and then describe the different design choices for visual encodings and interaction mechanisms that were made to fulfill the identified tasks. 

\subsection{Task Analysis}
\label{sec:tasks}
The task analysis was performed in collaboration with a domain expert in the field of HCI who developed, implemented, and executed the R2G experiments described in Section~\ref{sec:data_collection}. The outcome of the task analysis can be summarized as follows: \\
\textbf{\task{1}}: \emph{Global accuracy analysis for prediction models}. A global understanding of the accuracy of the classification outcome is desired to observe which accuracy can be achieved and whether (and how) the accuracy improves with increasing length of the time window used for training the model. \\
\textbf{\task{2}}: \emph{Identifying confusing classes}. For each of the trained models, it shall be possible to investigate which classes were predicted correctly and which classes were confused (leading to misclassifications). \\
\textbf{\task{3}}: \emph{Detecting trade-off point between prediction earliness and accuracy}. The identification of the model with the best trade-off between how early a prediction can be made and how accurate that prediction is shall be supported.\\
\textbf{\task{4}}: \emph{Detailed investigation of individual multi-variate time series}. For an in-depth analysis of individual multi-variate time series, it shall be possible to observe the prediction evolution of all classes over all models (i.e., models with increasing time window lengths). This is especially interesting when investigating time series with initial confusion to observe when the confusion was eliminated and when the predictions become reliable. \\
\textbf{\task{5}}: \emph{Impact of features.} Since the multivariate time series data consist of multiple features, it shall be possible to investigate, which features allow for distinguishing which classes for a selected model. This will lead to an understanding of which features are most relevant for the classification prediction. \\
%\textbf{\task{6}}: \emph{Evaluating model generalizability} — Assess the model's generalizability to new participants by performing a leave-one-out evaluation.\\

\subsection{Visual Design}
\begin{figure*}[htb]
    \centering
    \includegraphics[width=1\linewidth]{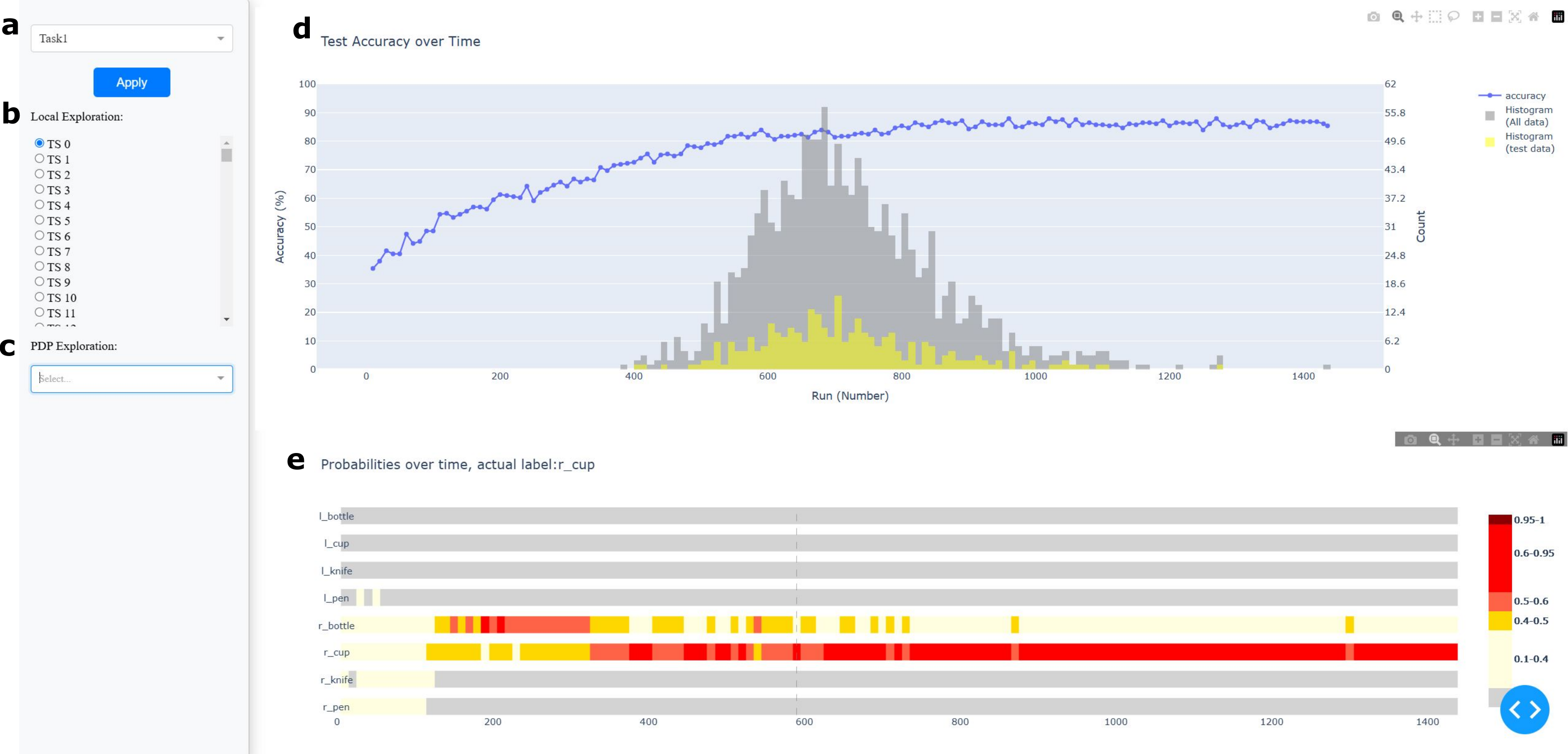}
    \caption{Overview of the XMTC tool, showing the user interface elements of the settings panel (a, b, c) and visualizations using coordinated views (d, e). The accuracy plot (d) illustrates the performance of prediction models with increasing time window (blue), accompanied by a histogram representing the distribution of the time series lengths of the entire dataset (gray) and the test dataset (yellow). Users can select a dataset in the settings panel (a) to explore model performance over time (d). Additionally, users can analyze individual test time series by selecting them in panel (b) and observing the models' class probabilities over time (e).}
    \label{fig:tool_overview_page1}
\end{figure*}

\begin{figure*}[htb]
    \centering
    \includegraphics[width=1\linewidth]{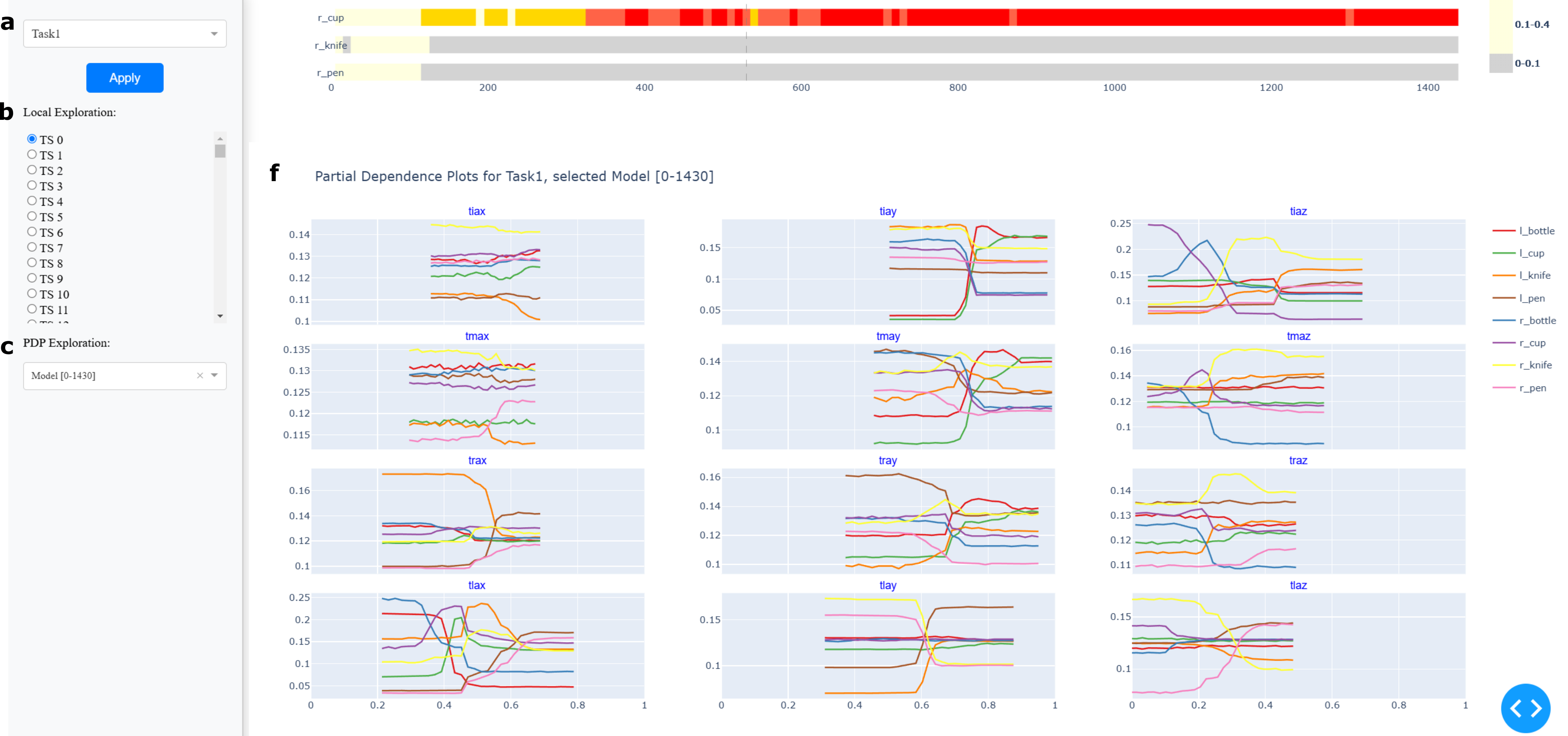}
    \caption{Overview of XMTC tool (continuation of Figure~\ref{fig:tool_overview_page1}). Users can select a model in the setting panel(c) and observe the Partial Dependence Plot (f) to find distinguishable features and patterns.}
    \label{fig:tool_overview_page2}
\end{figure*}

Guided by the outcome of the task analysis, we developed our XMTC approach for the analysis of multivariate time series classification predictions. We designed four visual encodings to address Tasks \task{1} - \task{5}. They are arranged in coordinated views and pop-up windows, as detailed below. Figures~\ref{fig:tool_overview_page1} and~\ref{fig:tool_overview_page2} provide an overview of the developed tool.

%The visual design of the XMTC tool is composed of multiple views, as illustrated in Figures~\ref{fig:tool_overview_page1} and~\ref{fig:tool_overview_page2}. The setting panel is located to the left which has three setting components(Figure~\ref{fig:tool_overview_page1}a, ~\ref{fig:tool_overview_page1}b, ~\ref{fig:tool_overview_page1}c).

\smallskip
\noindent
\textbf{Accuracy plot.} 
The accuracy plot shown in Figure~\ref{fig:tool_overview_page1}d provides an overview of model performance over time (Task~\task{1}).
It plots the accuracy of all models (y-axis) over the lengths of the time series datasets that were used to train the models (x-axis). It is shown as a blue curve with step size of 10 time steps.
Additionally, the plot provides histograms of the lengths of the collected time series of the entire dataset (grey) and the test dataset (yellow). 
The histograms allow for the detection of interesting time window sizes.

\smallskip
\noindent
\textbf{Confusion matrix.} To identify which classes are confused and how often (Task~\task{2}), we provide the confusion matrix. The square matrix plots the true class labels (y-axis) vs. the predicted class labels (x-axis) using a heatmap. For the heatmap we selected a multi-hue color map (purple-blue-green-yellow) with increasing brightness. In addition, the numbers (counts) are also provided in the matrix. In the case of a perfect classification, there would only be non-zero values on the diagonal. Any non-zero value of an entry that is not on the diagonal indicates a confusion (misclassification).

\begin{figure}[h!]
    \centering
    \includegraphics[width=1\linewidth]{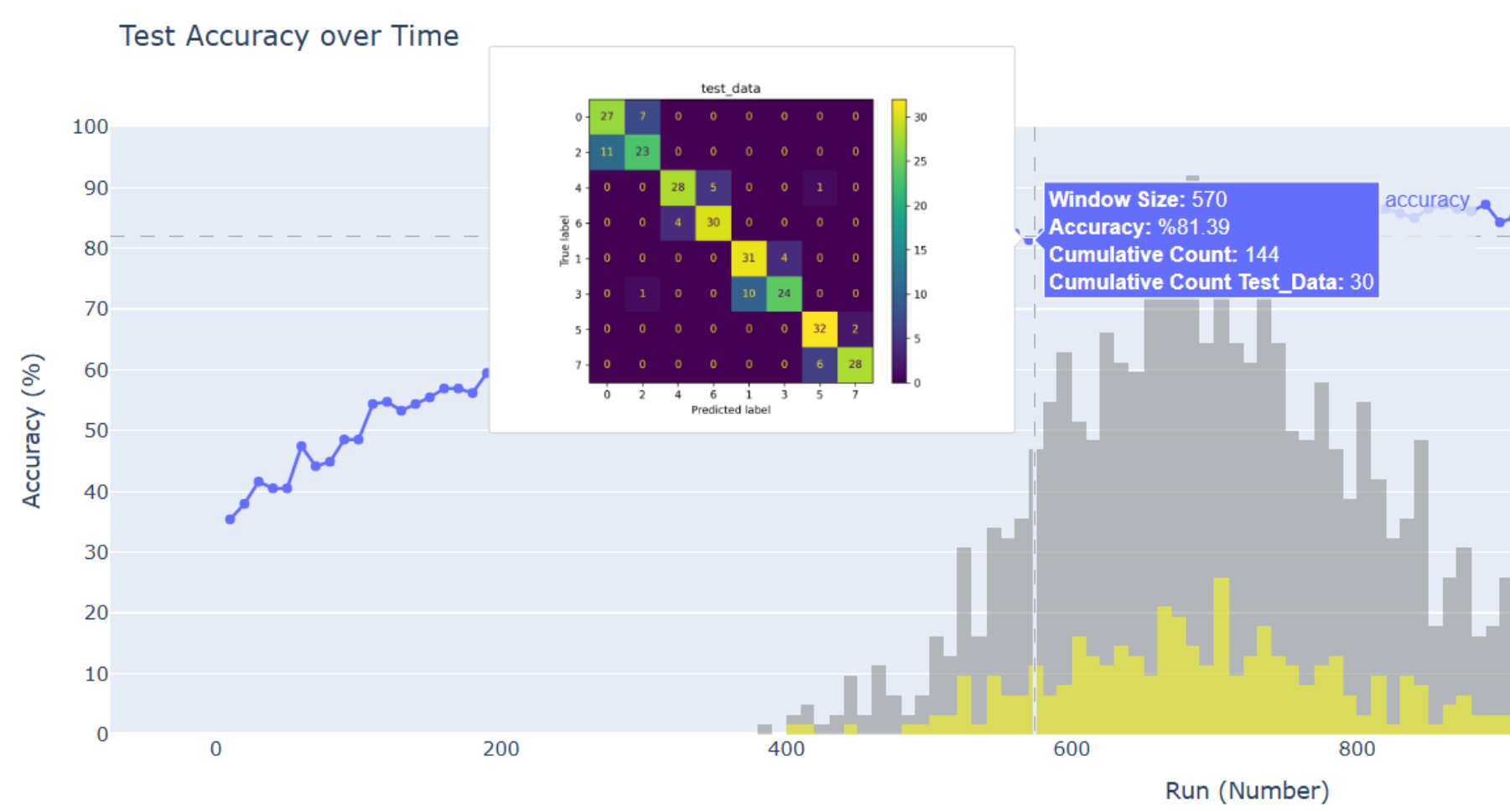}
    \caption{Hovering information of the accuracy plot. When the user hovers over the accuracy plot, details about the hovered model are displayed, including its confusion matrix visualization, window size, accuracy, the number of time series in the entire dataset shorter than the hovered window size, and the number of time series in the test dataset shorter than the window size.}
    \label{fig:tool_hover_accuracy}
\end{figure}

The confusion matrix can be shown for each model. This is triggered by hovering over the points of the blue curve in the accuracy plot. The confusion matrix for the model of the point that is hovered over is then shown in a pop-up window. Figure~\ref{fig:tool_hover_accuracy} shows such a confusion matrix visualization. In addition to the confusion matrix, further details about the selected model are provided as text, which includes the window size, the accuracy of the test dataset, the number of time series shorter than the hovered window size (counting the number of time series in the cumulative histogram of the entire dataset), and the number of time series in the test dataset is shorter than the window size (counting the number of time series in the cumulative histogram of the test dataset).
%over time (Task~\task{2}), the user can hover over points (representing models) in the accuracy plot to examine the confusion matrix for each model at different time steps. A confusion matrix is a straightforward powerful tool that evaluates the performance of a classification model by comparing its predictions to the actual class labels. It is a square table with an equal number of rows and columns corresponding to the number of class labels. The x-axis of the confusion matrix represents the predicted class labels, while the y-axis represents the true class labels. Each cell in the matrix contains the number of instances where a specific true label was predicted as a particular class, providing a detailed view of the model's classification performance, including areas of misclassification.
%hovering infos:
% When the user hovers over the accuracy plot in Figure~\ref{fig:tool_overview_page1}d, each point represents a DrCIF model. By hovering over a specific point (as shown in Figure~\ref{fig:tool_hover_accuracy}), the user can view detailed information about that model. This includes:  
% \begin{itemize}
%     \item The confusion matrix for the hovered model,
%     \item The window size,
%     \item The accuracy of the test dataset,
%     \item The number of time series shorter than the hovered window size (counting the number of time series in the cumulative histogram), and
%     \item The number of time series in the test dataset is shorter than the window size (counting the number of time series in the cumulative histogram of the test dataset).
% \end{itemize}

\smallskip
\noindent
\textbf{Time points.} 
The accuracy plot shows the evolution of the overall accuracy with increasing window size, while the confusion matrix provides an overview of which classes are predicted incorrectly and where the model exhibits confusion. As the user hovers over models at different points in time (windows sizes), the user can observe how this confusion changes over time.

Specific time steps offer valuable insights into model behavior and predictions:  
\begin{itemize}
    \item \textit{The initial time step}: Provides information on the earliest predictions.  
    \item \textit{The time step where none of the time series are over}: Indicates a stage all time series are still ongoing. (For the R2G data, this is the time point where the intended object has not yet been grasped by any participant.) This time step can be identified in the accuracy plot by investigating with histograms. 
    \item \textit{The time step after which there is no significant increase in the accuracy plot}: Highlights the point where model performance stabilizes. 
    \item \textit{The final time step}: Marks the point where all time series are complete.  
\end{itemize}  

According to the suggested time steps, the user can explore the models trained on the suggested window sizes, and explore and compare predictive models to identify the one that best classifies objects as early as possible (Task~\task{3}). The hovering information about the number of time series shorter than the hovered window size helps the user understand the number of time series that are already completed. This insight supports finding a balance between accuracy and earliness, allowing the user to evaluate the trade-offs effectively. 

\begin{figure*}[htb!]
    \centering
    \includegraphics[width=1\linewidth]{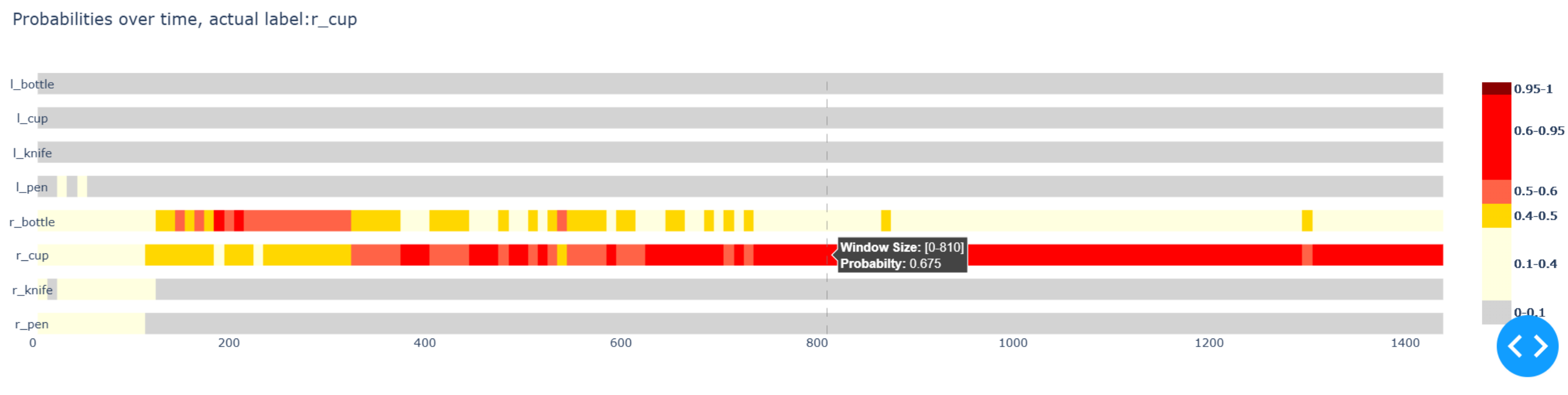}
    \caption{Hovering information of the heatmap: The window size and the probability predictions of the model trained on the corresponding window size are displayed and a vertical line is drawn to indicate the current time point.}
    \label{fig:tool_hover_heatmap}
\end{figure*}

\smallskip
\noindent
\textbf{Multivariate time series heatmap plot.}
To explore challenging and misclassified test time series and investigate the evolution of model predictions for individual test time series (Task~\task{4}), a heatmap plot for multivariate time series is provided. Upon selection of a specific test time series (in the setting panel, see Figure~\ref{fig:tool_overview_page1}b), the heatmap in Figure~\ref{fig:tool_overview_page1}e displays the models' probability predictions (inspired by~\cite{nourani2022detoxer}) over time. The title of the heatmap reveals the actual label of the selected time series. The heatmap's y-axis represents the dataset's class labels, while the x-axis captures the temporal progression, where the window size increases by 10. The heatmap values indicate the probability predictions of the selected model (x-axis) for each class (y-axis) for the chosen test time series. This visualization enables users to analyze how the models' predictions for the test time series align with or diverge from the true label, which includes the identification of any confusion with other classes over time. Thus, the user can examine the probability values assigned to each class over time and identify the time step after which the models become confident and predict the time series correctly.
The color map assigned to the heatmap is designed such that probabilities below $0.1$ are shown in grey, probabilities in the range $[0.1,0.5]$ are shown in shades of yellow (with decreasing brightness), and probabilities larger than $0.5$ are shown in shades of red (with decreasing brightness). Thus, probabilities higher than $50\%$ stand out. 
%
%hovering infos:
When the user hovers over the heatmap, the window size and the probability prediction of the model trained on that window size are displayed, see Figure~\ref{fig:tool_hover_heatmap}. Additionally, a vertical dashed line is shown to help the user compare the model's probability predictions across all labels.  

\smallskip
\noindent
\textbf{Partial Dependence Plot.}
To examine the global behavior of the model and identify distinguishable input features (Task~\task{5}), the Partial Dependence Plot for input features of the selected model is plotted. Partial Dependence Plots (PDPs) provide insights into how changes in a specific feature influence the model's predictions, helping users identify key features that drive classification decisions. This allows users to interpret the model's global behavior and detect potential biases or patterns in the data. 

To investigate PDPs for the test dataset, the user needs to select the model of interest (cf.~Figure~\ref{fig:tool_overview_page1}c) and the PDPs for the selected model are shown, see Figure~\ref{fig:tool_overview_page2}f. These PDPs are plotted for each feature and each class of the model.  
The PDP visualization uses small multiples, where the PDP of each feature is shown in a separate view using a juxtaposed layout.
In Figure~\ref{fig:tool_overview_page2}f, the small multiples are laid out such that each column of subplots represents the x-component of the corresponding feature (aperture). For example, the subplots in the first column display the x-components of apertures as \textit{tiax}, \textit{tmax}, \textit{trax}, and \textit{tlax}. Similarly, each row of subplots shows the x, y, and z components of the respective feature. For instance, the subplots in the first row represent \textit{tiax}, \textit{tiay}, and \textit{tiaz}, which correspond to the x, y, and z components of the Thumb-Index Aperture.  
In each PDP, the x-axis represents the value range of the respective feature, while the y-axis displays the model's predictions for each class, i.e., a curve is plotted for each class. Different colors are used for different classes and a color legend is provided. The user can select or deselect specific classes to compare their PDPs and identify distinguishing features between the selected classes. 

\smallskip
\noindent
\textbf{Analytical workflow.}
The user would start the analysis by loading a data set (cf.~Figure~\ref{fig:tool_overview_page1}a) and investigating the global accuracy evolution in the accuracy plot. The user can, then, investigate individual time points by hovering and triggering the confusion matrix visualizations. When identifying a confusion in the confusion matrix, the user can select individual multivariate time series and analyze their prediction evolution in detail. When identifying a specific model, the user can then use the partial dependency plots to investigate, which features allowed the model the distinguish the classes. Altogether, the user can then decide what would be a good trade-off between an early prediction and an accurate prediction.

\section{Case Studies}
In this section, we discuss the use cases and findings of the XMTC tool, using R2G hand motion tracking data for object prediction from three experiments (Experiments 1, 2, and 3) for which data was collected (as described in Section~\ref{sec:data_collection}). To address Tasks~\task{1}, \task{2}, and \task{3}, the accuracy plot together with the confusion matrix visualization serves as a key visualization, while the heatmap plot is particularly useful for Task~\task{4}, and the Partial Dependence Plot (PDP) provides valuable insights for Task~\task{5}.

\subsection{Analysis of Experiment 1}
\label{sec:task1_analysis}
For this experiment, we use a total of $1,370$ multivariate time series, with the longest series containing $1,436$ time steps. Each model is trained using $80\%$ of the data, while the remaining $20\%$ is reserved for testing. The training and testing datasets are randomly selected and stratified to ensure balance across classes. As described in Section~\ref{sec:model}, we applied DrCIF with its default settings for each window size over time.

\begin{figure*}[htb!]
    \centering
    \includegraphics[width=1\linewidth]{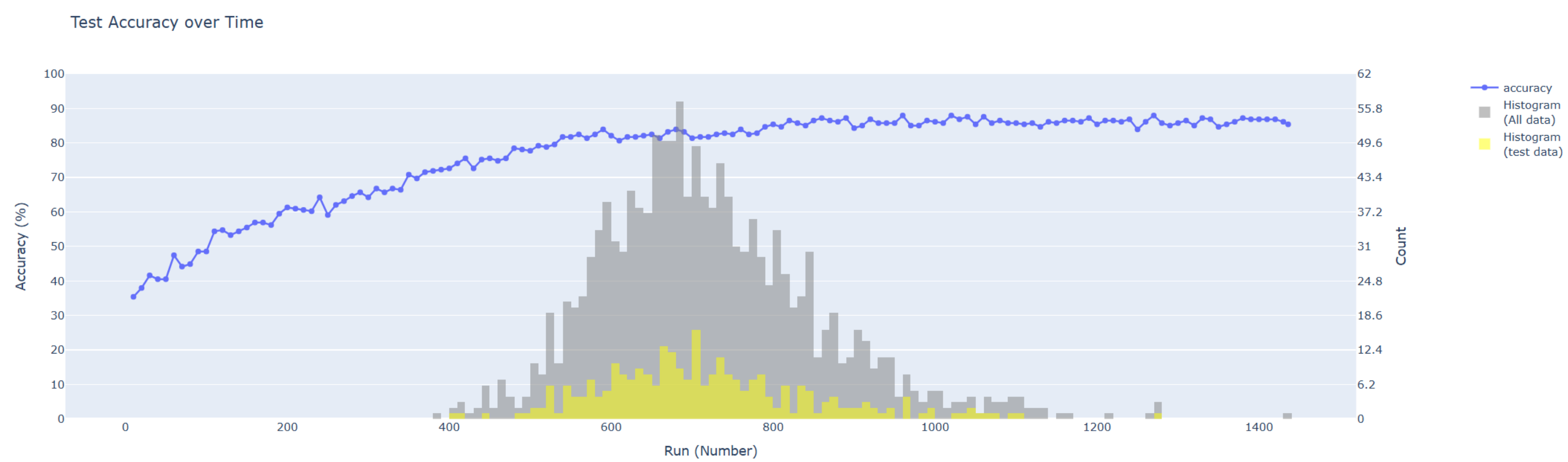}
    \caption{Accuracy plot of the test dataset for Experiment 1, starting at $35.4\%$ for the initial window size and reaching $85.4\%$ for the final window size. The histogram of the entire dataset (gray) indicates that most time series have lengths in the range of [600-800], which is consistent with the histogram for the test dataset (yellow).}
    \label{fig:Tool_Task1_accuracy}
\end{figure*}

\noindent
\textbf{Accuracy plot observation.}
When the user selects \textit{Experiment 1} in the settings panel, the accuracy plot is displayed (see Figure~\ref{fig:Tool_Task1_accuracy}).
In this experiment, we have eight conditions with four different objects located either on the left-hand or the right-hand side of the participant.  The goal of the classification is to detect these eight conditions, i.e., deciding on which object will be grasped and whether it is located on the left or the right. We note that the objects bottle and cup are quite similar to each other and so are the objects knife and pen.
The accuracy plot shows the models' accuracy, with the model for the smallest window size ([0,10]) achieving $35.4\%$ and for the largest window size ([0,1436]) reaching $85.4\%$. 

\begin{figure}[htb!]
    \centering
    \includegraphics[width=1\linewidth]{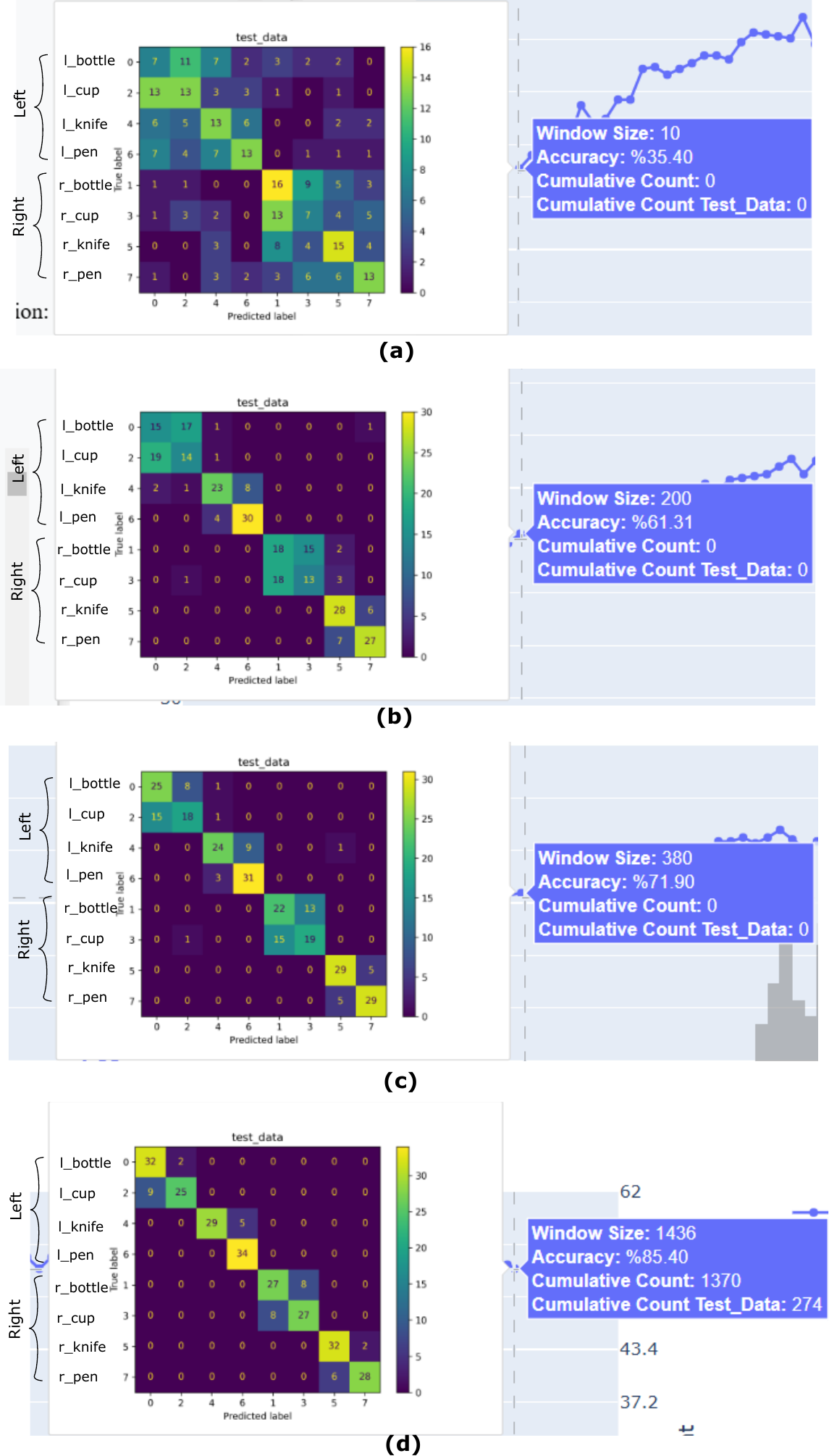}
    \caption{Confusion matrix and accuracy of the four models with window sizes from 0 to 10 (shortest window size), 200, 380, and 1436 (largest window size) for Experiment 1.}
    \label{fig:Tool_Task1_confusion_matrix_10_200_380_1436}
\end{figure}

\noindent
\textbf{Confusion matrix plot observation.}
By hovering over the points (representing models) and examining the confusion matrix for the smallest window size (Figure~\ref{fig:Tool_Task1_confusion_matrix_10_200_380_1436}a), we observe that classes corresponding to objects located on the left-hand side are primarily distinguishable from those located on the right side.
By hovering over the model with a window size of [0,200], we observe that the model's predictions improve ($61.31\%$). At this stage, objects are primarily confused only with similar ones (bottle with cup and knife with pen) in the same direction (left or right), as shown in Figure~\ref{fig:Tool_Task1_confusion_matrix_10_200_380_1436}b. %In Experiment 1, the bottle and cup are notably similar, as are the pen and knife.
Another interesting model is the model trained on the window size of [0,380] when all time series are shorter than this window size (meaning that none of the participants has grasped the intended object yet). This model has an accuracy of $71.9\%$, see Figure~\ref{fig:Tool_Task1_confusion_matrix_10_200_380_1436}c.
%
%comparison of Confusion matrix for 4 models with window sizes: 10, 200, 380, 1436
For the final time point (window size of [0,1436]), we observe that the confusions further decrease, 
see Figure~\ref{fig:Tool_Task1_confusion_matrix_10_200_380_1436}d.
%the model with the shortest window size, the model with a 200-time-step window size (where objects are only confused with similar ones in the same direction, left or right), the model with 380-time-step window size (which corresponds to the last time step before all time series are shorter than this duration), and the model with 1436-time-step window size (which is the longest model where all time series are complete).
% comparison of 4 models object wise
Table~\ref{table:Task1_comparison_4models} shows the respective numbers for the four time points mentioned above (called Model 10, 200, 380, and 1436, respectively). It shows that \textit{l\_pen} is improving very fast, achieving $88\%$ accuracy in Model 200 with \textit{l\_cup} following next, while \textit{r\_cup} and \textit{r\_bottle} are the most difficult objects to classify in all four models.

\begin{table*}[ht]
    \centering
    %\small
    \resizebox{\textwidth}{!}{%
    %\begin{tabular}{|l|c|c|c|c|c|c|c|c|c|}
    \begin{tabular}{|l|p{1.5cm}|p{1.5cm}|p{1.5cm}|p{1.5cm}|p{1.5cm}|p{1.5cm}|p{1.5cm}|p{1.5cm}|p{1.5cm}|}
        \hline
        \textbf{Class Label} & \textbf{Total Test Samples} & \textbf{Number of test samples correctly predicted (Model 10)} & \textbf{Number of test samples correctly predicted (Model 200)} & \textbf{Number of test samples correctly predicted (Model 380)} & \textbf{Number of test samples correctly predicted (Model 1436)} & \textbf{Accuracy (Model 10) (\%)} & \textbf{Accuracy (Model 200) (\%)} & \textbf{Accuracy (Model 380) (\%)} & \textbf{Accuracy (Model 1436) (\%)} \\ \hline
        l\_bottle    &34     & 7     & 15    &25     &32     &21\% 	    &44\%     &74\%     &94\%\\ \hline
        l\_cup       &24     & 13    & 14    &18     &25     &38\%     &41\%     &53\%     &74\%\\ \hline
        l\_knife     &34     & 13    & 23    &24     &29     &38\%     &68\%     &71\%     &85\%\\ \hline
        l\_pen       &34     & 13    & 30    &31     &34     &38\%     &88\%     &91\%     &100\%\\ \hline
        r\_bottle    &35     & 16    & 18    &22     &27     &46\%     &51\%     &63\%     &77\%\\ \hline
        r\_cup       &35     & 7     & 13    &19     &27     &20\%     &37\%     &54\%     &77\%\\ \hline
        r\_knife     &34     & 15    & 28    &29     &32     &44\%     &82\%     &85\%     &94\%\\ \hline
        r\_pen       &34     & 13    & 27    &29     &28     &38\%     &79\%     &85\%     &82\%\\ \hline
    \end{tabular}
    }
    \caption{Comparison of models' performance across different window sizes for Experiment 1.}
    \label{table:Task1_comparison_4models}
\end{table*}

\begin{figure*}[htb!]
    \centering
    \includegraphics[width=1\linewidth]{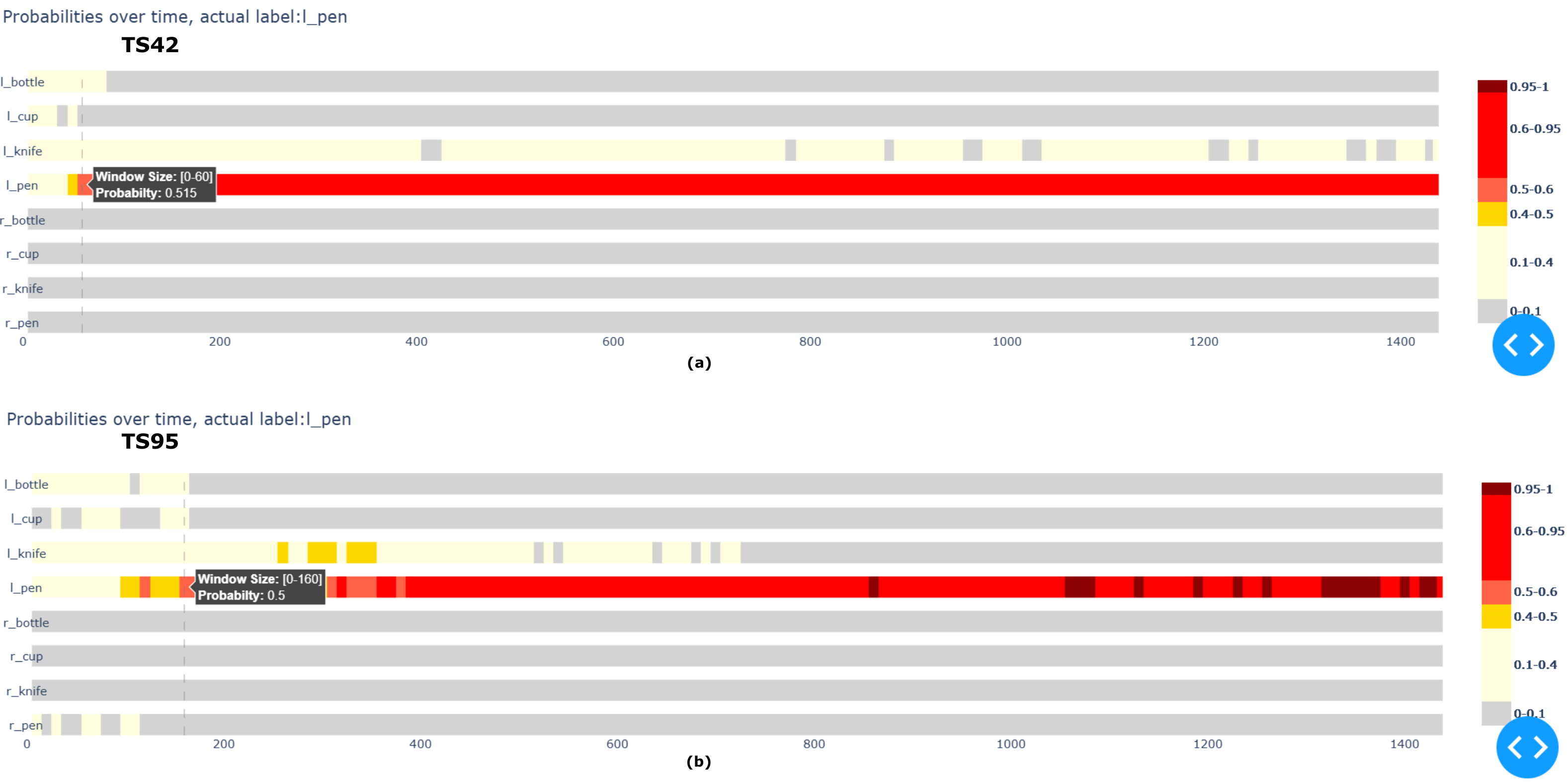}
    \caption{Models' confidence over time for two samples in the \textit{l\_pen} class. All models with a window size greater than 60 predict the first sample (a) with high confidence, assigning a probability above 0.6. Similarly, all models with a window size greater than 160 predict the second sample (b) with confidence above 0.5, reaching over 0.95 for some window sizes.}
    \label{fig:Tool_Task1_heatmap_l_pen}
\end{figure*}

\noindent
\textbf{Heatmap plot observation.}
After exploring the accuracy and confusion matrix plots of models over time, the user can examine the test time series by selecting the desired series in the settings panel (Figure~\ref{fig:tool_overview_page1}b). When analyzing the accuracy plot for Experiment 1, we observed that \textit{l\_pen} is the most distinguishing class. To further investigate the confidence of the models' predictions over time, two time series from this class were selected and visualized in Figure~\ref{fig:Tool_Task1_heatmap_l_pen}. The red regions in the plot indicate predictions with confidence above 0.5. For the selected time series, the models demonstrate consistently high confidence in predicting \textit{l\_pen} over time.

\begin{figure*}[htb]
    \centering
    \includegraphics[width=1\linewidth]{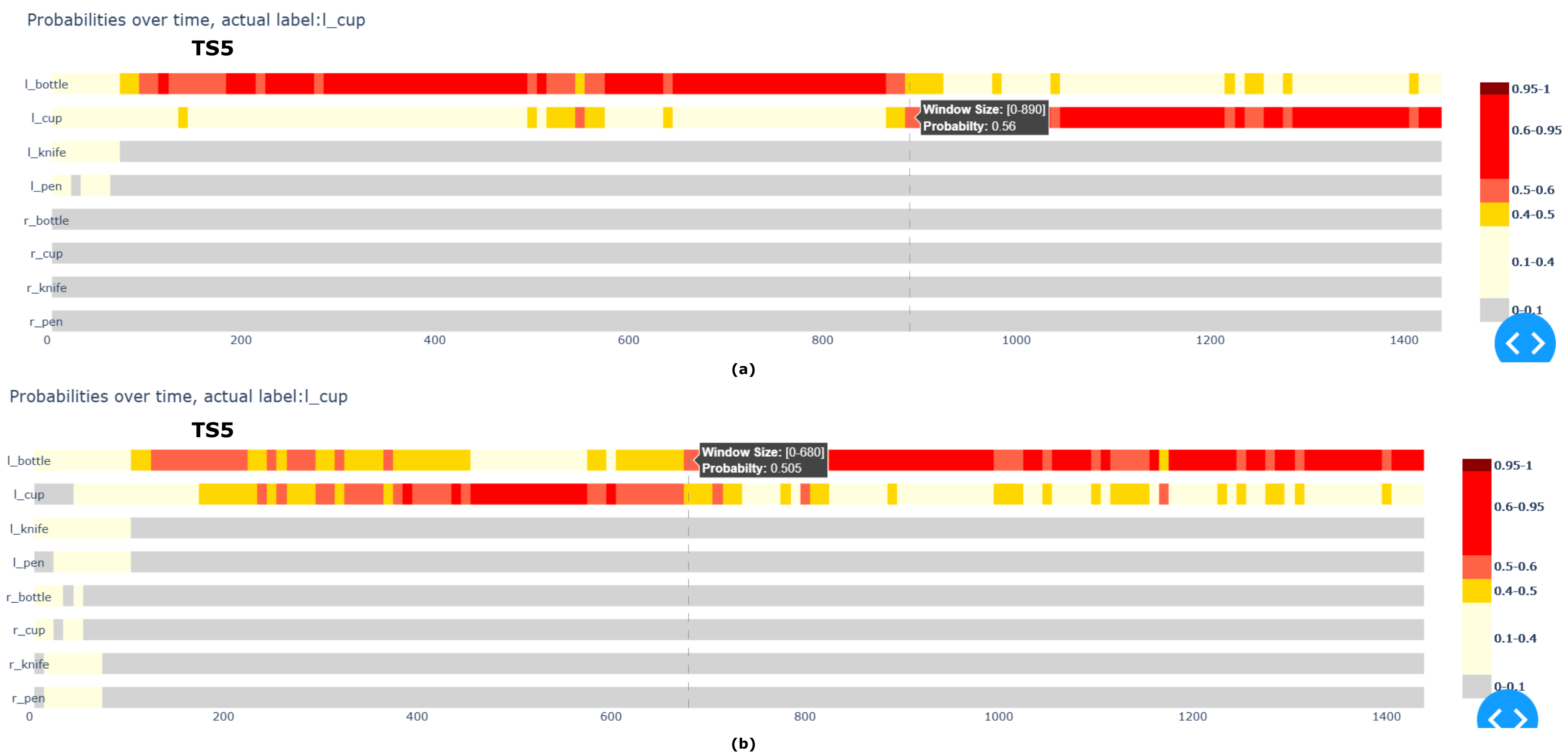}
    \caption{Models' confidence over time for two samples in the \textit{l\_cup} class, the most challenging class for object prediction. The first sample (a) is misclassified as \textit{l\_bottle} until a window size of 890, after which models with larger window sizes predict it correctly. The second sample (b) is another difficult case, also confused with \textit{l\_bottle}, and even at larger window sizes of 680, it remains misclassified as \textit{l\_bottle}.}
    \label{fig:Tool_Task1_heatmap_l_cup}
\end{figure*}

On the other hand, class \textit{l\_cup} was the most challenging class to predict accurately. To analyze this further, we plotted two time series from this class in Figure~\ref{fig:Tool_Task1_heatmap_l_cup}. The first time series in Figure~\ref{fig:Tool_Task1_heatmap_l_cup}a, reveals that models consistently misclassified it as class \textit{l\_bottle} before time step 890. In the second example, depicted in Figure~\ref{fig:Tool_Task1_heatmap_l_cup}b, all models show low confidence in their predictions. Initially, the models predict the time series as \textit{l\_bottle}, followed by a brief prediction as \textit{l\_cup}, and eventually reverted to \textit{l\_bottle}, which is the incorrect label.

\begin{figure*}
    \centering
    \includegraphics[width=1\linewidth]{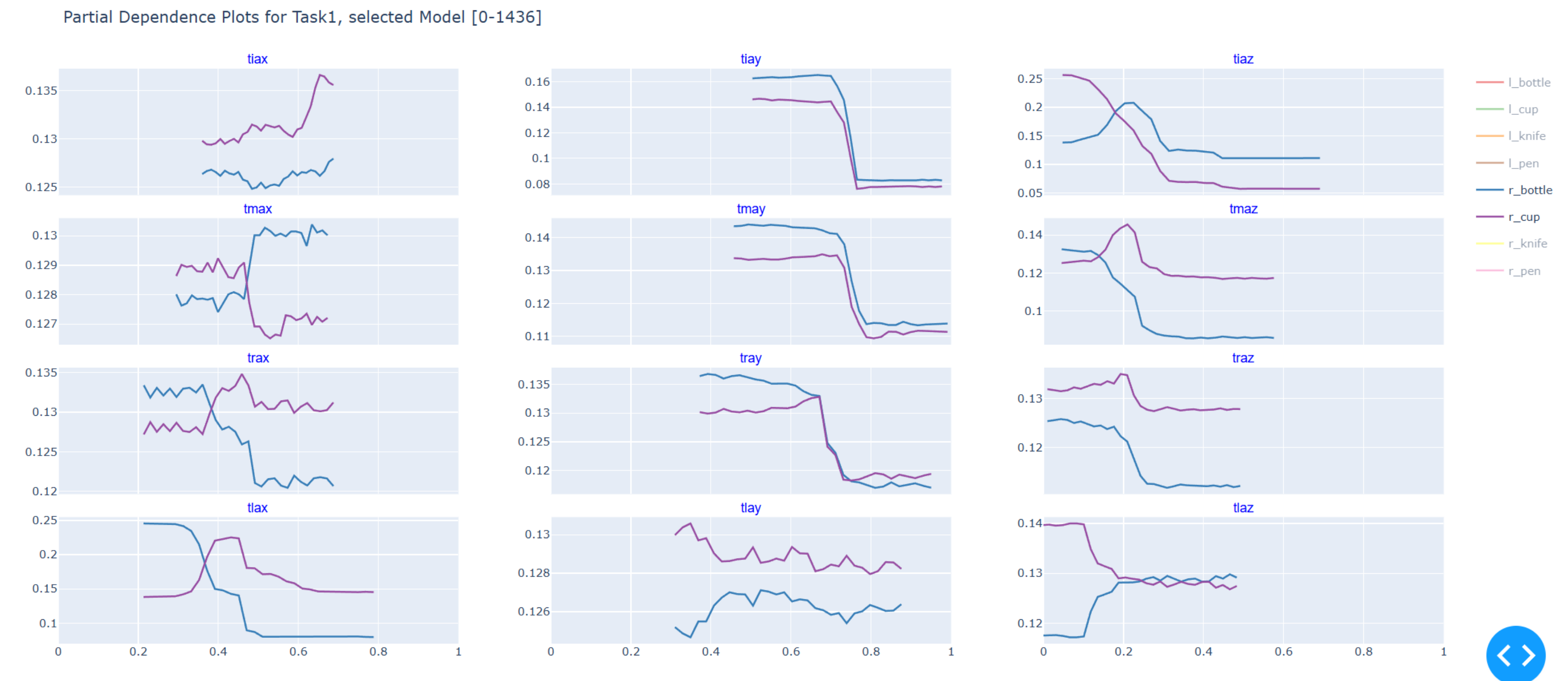}
    \caption{Partial Dependence Plot for two challenging classes, \textit{r\_cup} and \textit{r\_knife}, in Experiment 1. Features such as \textit{tiay}, \textit{tmay}, and \textit{tray} exhibit similar patterns, making them ineffective for distinguishing between these two classes. In contrast, features like \textit{tmax}, \textit{trax}, and \textit{tlax} display more distinctive patterns, providing better class separation.}
    \label{fig:Tool_Task1_PDP_r_cup_r_bottle}
\end{figure*}

\noindent
\textbf{Partial Dependence Plot observation.}
 To identify distinguishing and non-distinguishing features, the user can select a specific model in the settings panel (Figure~\ref{fig:tool_overview_page1}c), which generates the Partial Dependence Plot for the selected model. Based on our analysis, \textit{r\_cup} and \textit{r\_knife} were identified as the most challenging classes for object prediction. The PDPs for these classes are shown in Figure~\ref{fig:Tool_Task1_PDP_r_cup_r_bottle}.
The PDPs highlight that features such as \textit{tiay}, \textit{tmay}, and \textit{tray} exhibit similar patterns, making them less effective for distinguishing between the two classes. In contrast, features like \textit{tmax}, \textit{trax}, and \textit{tlax} demonstrate more distinctive patterns. For instance, the model's prediction for the \textit{r\_cup} class shows a decreasing trend, whereas the prediction for the \textit{r\_bottle} class exhibits an increasing trend, indicating that the features in the x-direction have the potential for improving classification accuracy.

\subsection{Analysis of Experiment 2}
The total number of time series in this experiment is $1,388$, where the longest time series has a length of $2,842$. Similar to Experiment 1, each model is trained using $80\%$ of the data, while the remaining $20\%$ is reserved for testing. 

\begin{figure*}[htb]
    \centering
    \includegraphics[width=1\linewidth]{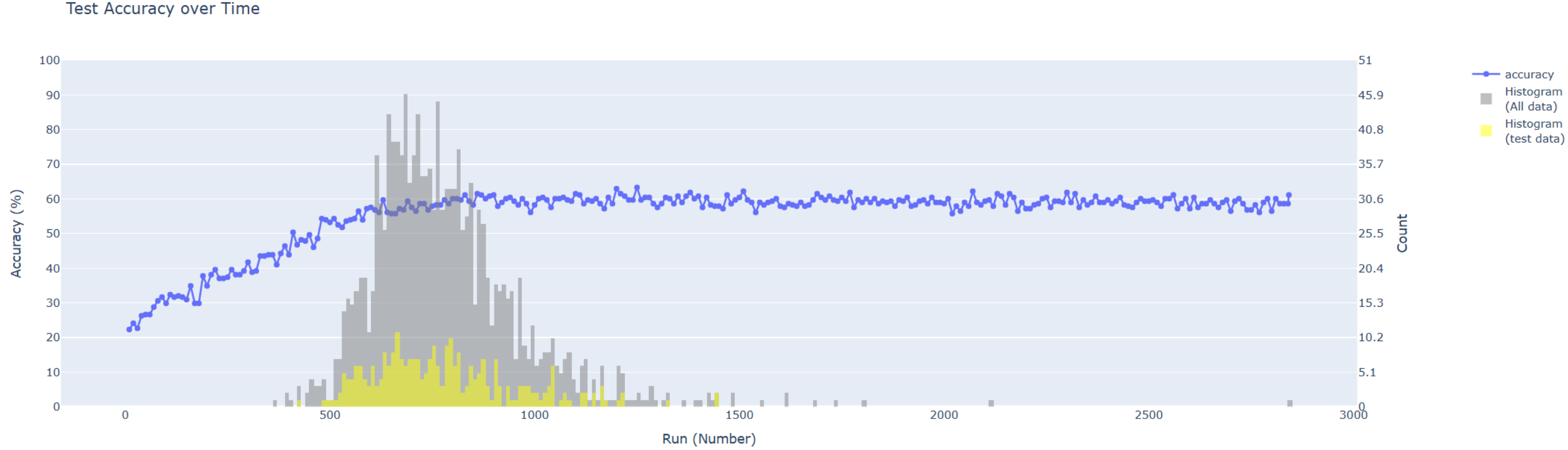}
    \caption{Accuracy of test dataset for Experiment 2. Achieving an initial accuracy of $22.30\%$ and increasing to $59.71\%$ by time step 630, stabilizing at approximately $60\%$ for the remaining time steps which shows that Experiment 2 is not performing as expected for object prediction.}
    \label{fig:Tool_Task2_accuracy}
\end{figure*}

\noindent
\textbf{Accuracy plot observation.}
Figure~\ref{fig:Tool_Task2_accuracy} illustrates the accuracy of the models over time. The model with the shortest window size ([0,10]) achieves an initial accuracy of $22.30\%$, which increases to $59.71\%$ by time step 630, stabilizing at approximately $60\%$ for the remaining time steps. This indicates that the object classification for Experiment 2 is not performing as expected.
Following discussions with the expert responsible for data collection, we determined that, in Experiment 2, participants were instructed to \textit{grasp}, \textit{touch}, or \textit{push} objects without focusing on the specific object types. This aligns with our observation that none of the models achieve high performance over time, supporting the expert's statement. 
As a result, we do not proceed with other analysis tasks for this experiment.

\subsection{Analysis of Experiment 3}
In this experiment, the total number of time series is $1,834$ and the longest one has a length of $1,378$. Similar to Experiments 1 and 2, each model is trained using $80\%$ of the data, while the remaining $20\%$ is reserved for testing. 

\begin{figure*}[htb]
    \centering
    \includegraphics[width=1\linewidth]{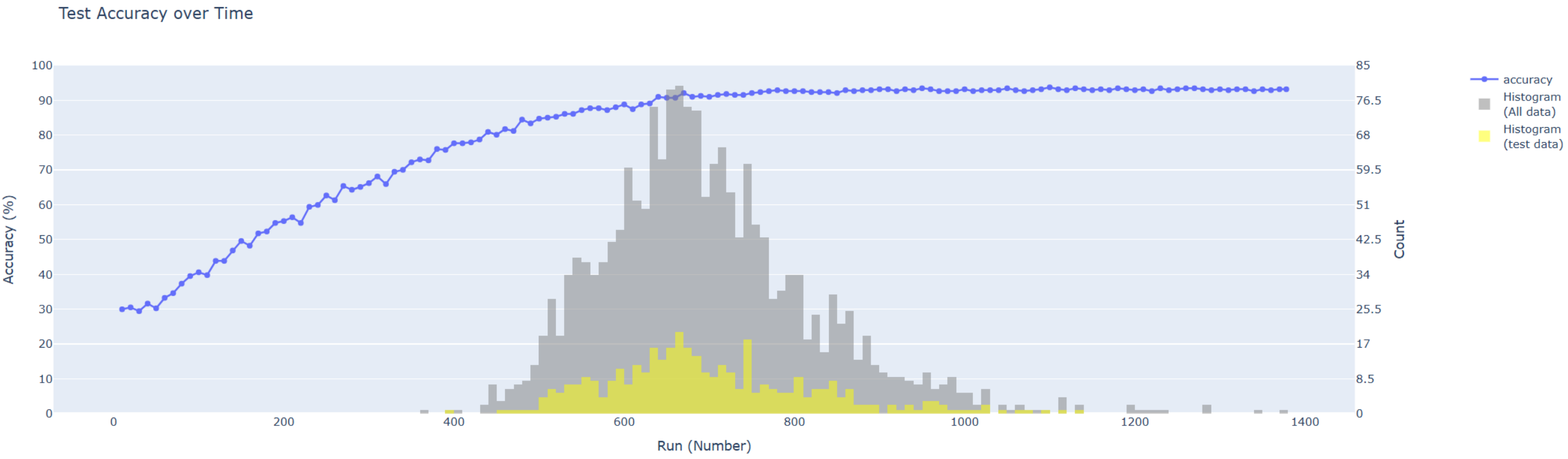}
    \caption{Accuracy plot of the test dataset for Experiment 3. The model initially achieves an accuracy of $29.97\%$, which increases to $73.02\%$ at time step 360 and reaches $91.01\%$ by time step 640. After time step 640, the accuracy stabilizes, and at the final time step (1378), the model attains an accuracy of $93.19\%$.}
    \label{fig:Tool_Task3_accuracy}
\end{figure*}

\begin{figure}[htb!]
    \centering
    \includegraphics[width=1\linewidth]{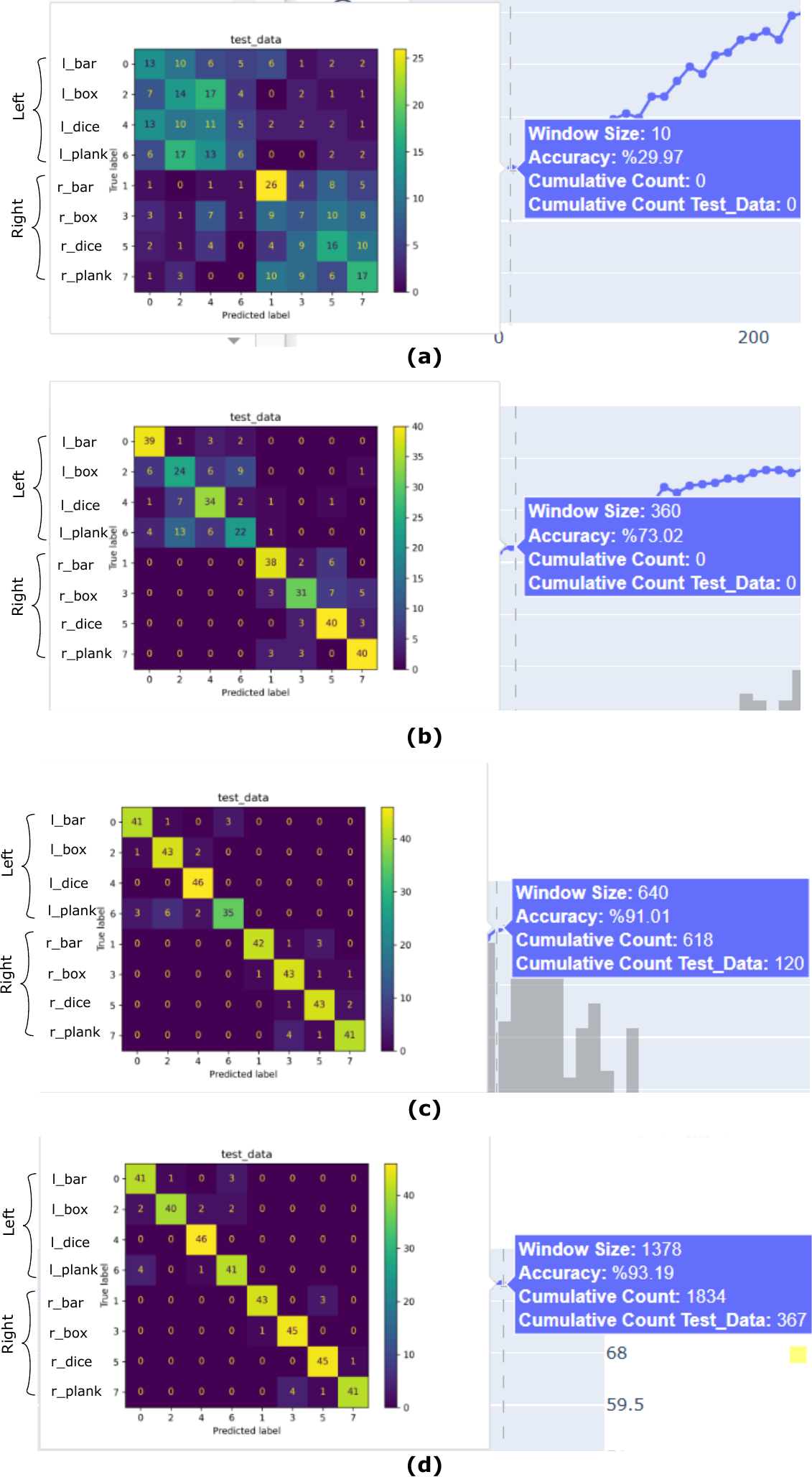}
    \caption{Confusion matrix and accuracy of the four models with window sizes of 10 (shortest window size), 360, 640, and 1378 (largest window size) for Experiment 3.}
    \label{fig:Tool_Task3_confusion_matrix_10_360_640_1378}
\end{figure}

\noindent
\textbf{Accuracy plot observation.}
Figure~\ref{fig:Tool_Task3_accuracy} shows the accuracy plot for Experiment 3 over time.  
For this Experiment, the most notable time steps include the initial time step (window size=10), the last time step before any object is grasped (window size=360), the time step after which the accuracy remains relatively steady (window size=640), and the final time step (window size=1378). 
Initially, the model achieves an accuracy of $29.97\%$, which increases to $73.02\%$ at time step 360 and reaches $91.01\%$ by time step 640. The accuracy remains quite steady after time step 640 and at the final time step (1378), the model achieves an accuracy of $93.19\%$ (see Figure~\ref{fig:Tool_Task3_confusion_matrix_10_360_640_1378}). 
If the user prioritizes earliness, the plot suggests selecting the model with a window size of 360 (see Figure~\ref{fig:Tool_Task3_confusion_matrix_10_360_640_1378}b), which predicts the intended object before it is grasped (i.e., before the user touches the object) with an accuracy of $73.02\%$. However, if the user prioritizes higher accuracy, even if it requires allowing some objects to be grasped, the model with a window size of 640 is recommended, as it achieves a steady accuracy thereafter.

%comparison of Confusion matrix for 4 models with window sizes: 10, 360, 640, 1378
\noindent
\textbf{Confusion matrix plot observation.}
For further analysis of these four time steps for each object, Figure~\ref{fig:Tool_Task3_confusion_matrix_10_360_640_1378} shows the confusion matrix visualizations and Table~\ref{table:Task3_comparison_4models} presents the results, referring to the time steps as Model 10, Model 360, Model 640, and Model 1378. In Model 10, \textit{r\_bar} is predicted with high accuracy compared to the other objects, with only 10 time steps. Another key observation is that \textit{l\_dice} achieves 100\% accuracy for Model 640 and maintains this accuracy thereafter. For \textit{l\_bar} and \textit{r\_plank}, the improvement from Model 360 to Model 640 is minimal, at 4\% and 2\%, respectively (i.e., 2 and 1 additional correct predictions). Furthermore, no improvement is observed from Model 640 to Model 1378.

\begin{table*}[ht]
    \centering
    %\small
    \resizebox{\textwidth}{!}{%
    %\begin{tabular}{|l|c|c|c|c|c|c|c|c|c|}
    \begin{tabular}{|l|p{1.5cm}|p{1.5cm}|p{1.5cm}|p{1.5cm}|p{1.5cm}|p{1.5cm}|p{1.5cm}|p{1.5cm}|p{1.5cm}|}
        \hline
        \textbf{Class Label} & \textbf{Total Test Samples} & \textbf{Number of test samples correctly predicted (Model 10)} & \textbf{Number of test samples correctly predicted (Model 360)} & \textbf{Number of test samples correctly predicted (Model 640)} & \textbf{Number of test samples correctly predicted (Model 1378)} & \textbf{Accuracy (Model 10) (\%)} & \textbf{Accuracy (Model 360) (\%)} & \textbf{Accuracy (Model 640) (\%)} & \textbf{Accuracy (Model 1378) (\%)} \\ \hline
        l\_bar      &45     & 13    & 39    &41     &41     &29\% 	  &87\%     &91\%     &91\%\\ \hline
        l\_box      &46     & 14    & 24    &43     &40     &30\%     &52\%     &93\%     &87\%\\ \hline
        l\_dice     &46     & 11    & 34    &46     &46     &24\%     &74\%     &100\%    &100\%\\ \hline
        l\_plank    &46     & 6     & 22    &35     &41     &13\%     &48\%     &76\%     &89\%\\ \hline
        r\_bar      &46     & 26    & 38    &42     &43     &57\%     &83\%     &91\%     &93\%\\ \hline
        r\_box      &46     & 7     & 31    &43     &45     &15\%     &67\%     &93\%     &98\%\\ \hline
        r\_dice     &46     & 16    & 40    &43     &45     &35\%     &87\%     &93\%     &98\%\\ \hline
        r\_plank    &46     & 17    & 40    &41     &41     &37\%     &87\%     &89\%     &89\%\\ \hline
    \end{tabular}
    }
    \caption{Comparison of model performance across different window sizes for Experiment3.}
    \label{table:Task3_comparison_4models}
\end{table*}

\begin{figure*}[htb]
    \centering
    \includegraphics[width=1\linewidth]{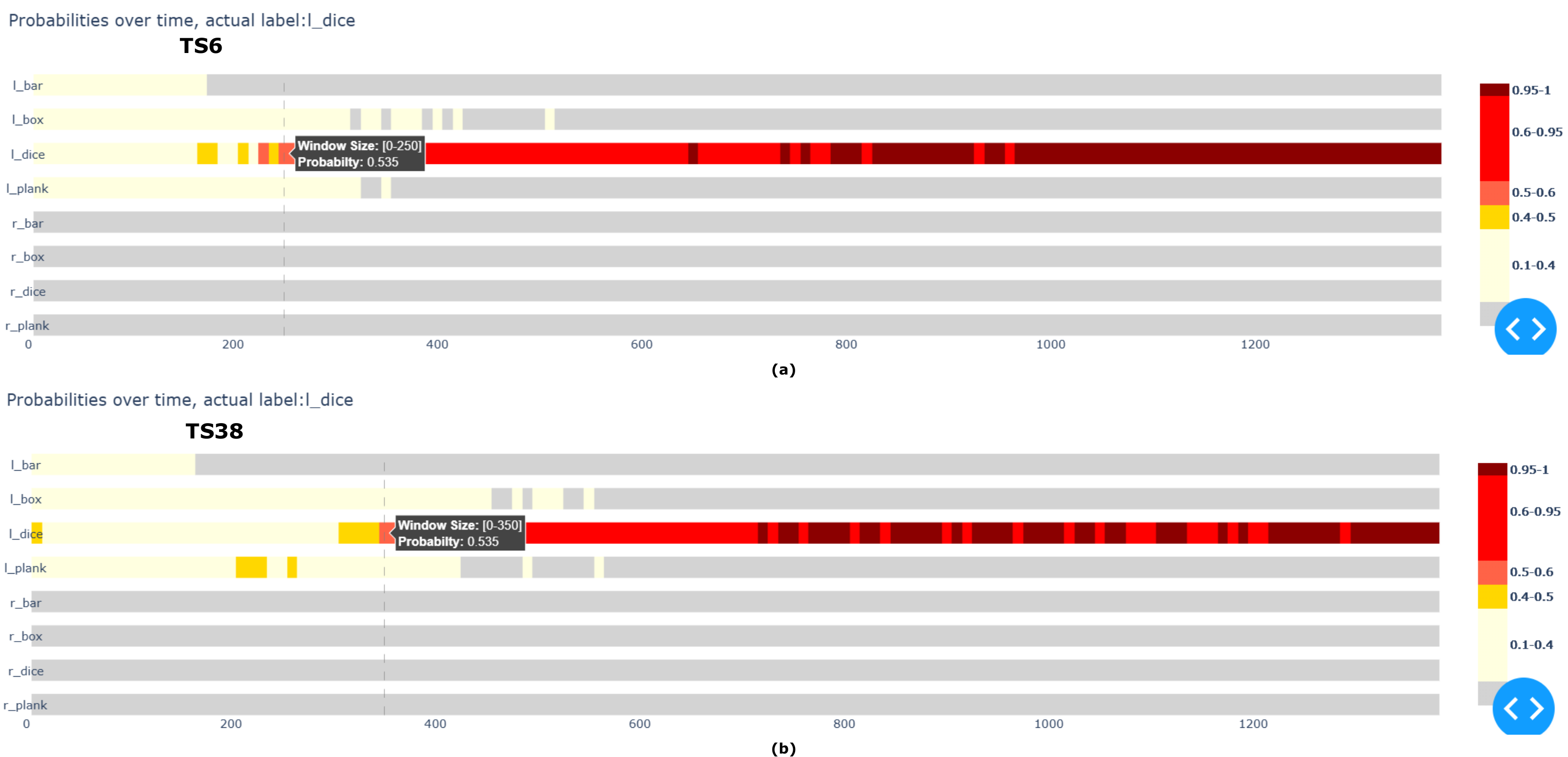}
    \caption{Model's confidence over time for two samples in the \textit{l\_dice} class in Experiment 3. The first sample (a) is predicted with high confidence from a window size of 250, reaching a probability above 0.6, and for most window sizes larger than 800, the confidence exceeds 0.95. The second sample (b) is correctly predicted with confidence above 0.6 from a window size of 350, and for many window sizes larger than 800, the probability surpasses 0.95.}
    \label{fig:Tool_Task3_heatmap_l_dice}
\end{figure*}

\noindent
\textbf{Heatmap plot observation.}
To explore model predictions and confidence about how the time series are confused with other classes over time, the heatmap plot can be explored.
In Figure~\ref{fig:Tool_Task3_heatmap_l_dice}, two random time series from the class \textit{l\_dice} are selected to observe the model prediction probabilities over time. According to Table~\ref{table:Task3_comparison_4models}, models with a window size of at least 640 are able to predict the class label of this object with 100\% accuracy. In Figure~\ref{fig:Tool_Task3_heatmap_l_dice}, the time series' classes are correctly predicted with high confidence after time steps 250 and 350, respectively, providing further support for our findings.

\begin{figure}[htb!]
    \centering
    \includegraphics[width=1\linewidth]{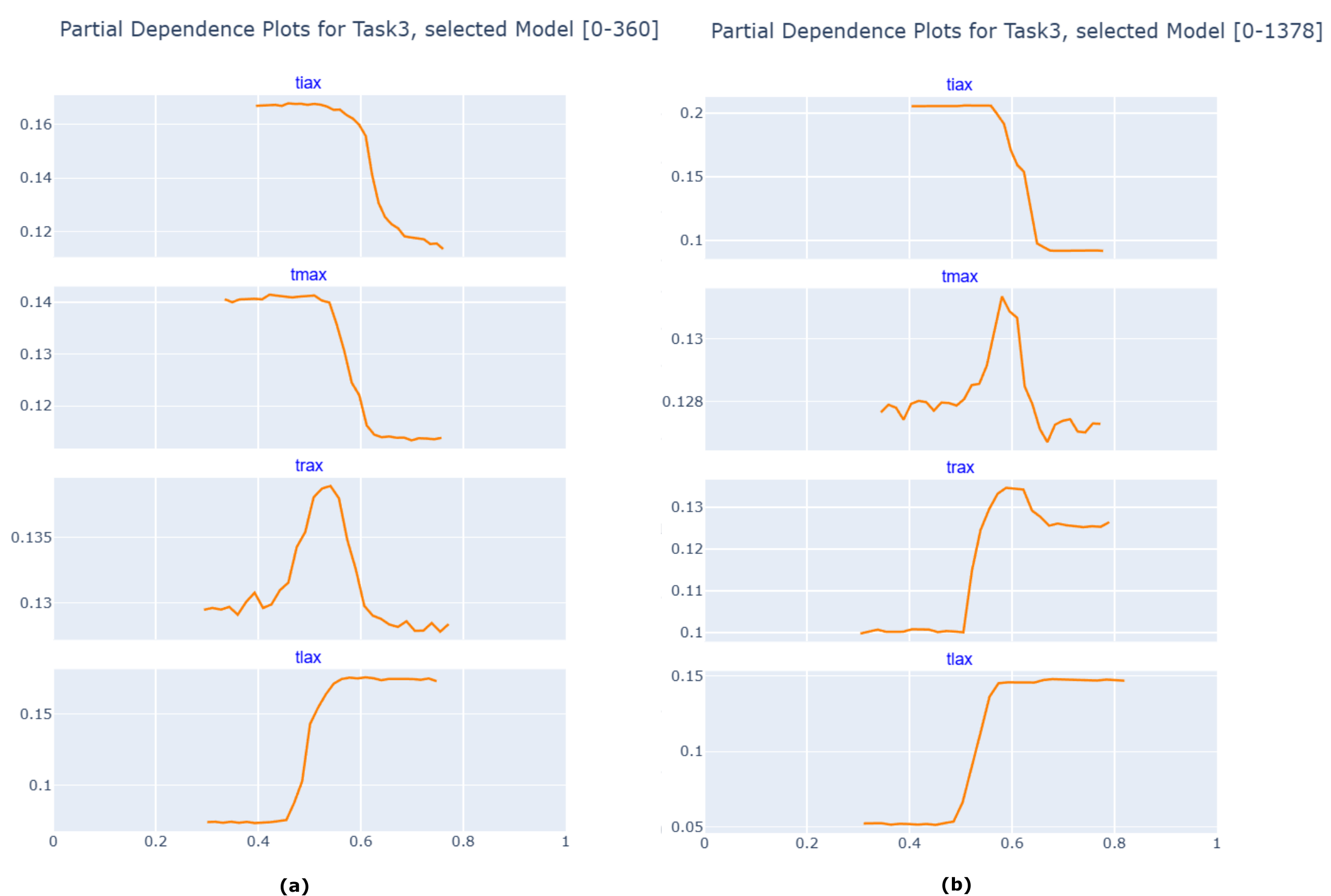}
    \caption{Partial Dependence Plots of the \textit{l\_dice} class for two time steps (360 and 1378) in Experiment 3. The model's accuracy at these time steps is 74\% and 100\%, respectively. The features \textit{tiax} and \textit{tlax} exhibit very similar patterns, while \textit{tmax} and \textit{trax} show noticeable changes. This indicates that the x-components of Thumb-Middle (\textit{tmax}) and Thumb-Ring (\textit{trax}) Apertures play a critical role in improving prediction performance for \textit{l\_dice}.}
    \label{fig:Tool_Task3_PDP_l_dice_360_1378}
\end{figure}

\noindent
\textbf{Partial Dependence Plot observation.}
To identify the distinguishing input features over time for the class \textit{l\_dice}, which achieves 100\% accuracy, we analyze the PDP of this class in Figure~\ref{fig:Tool_Task3_PDP_l_dice_360_1378}. The comparison is made between the model at time step 360, which predicts this class with 74\% accuracy, and the final model, which achieves 100\% accuracy. In this plot, the features \textit{tiax} and \textit{tlax} exhibit very similar patterns, while \textit{tmax} and \textit{trax} show noticeable changes. This suggests that the x-components of Thumb-Middle and Thumb-Ring Apertures play a critical role in improving the prediction performance for \textit{l\_dice}, increasing accuracy from 74\% at time step 360 to 100\% at the final time step.

\subsection{Leave-one-out Test}
When analyzing the results above with the HCI domain expert, he pointed out that he would be interested in assessing the generalizability of our classification to unseen users. Thus, we conducted a leave-one-out test on Experiment 1. In this evaluation, the model was trained on time series data from all but one user, and the left-out user's data was used for testing. This process was repeated for each user and for each window size in Experiment 1.

\begin{figure*}[htb]
    \centering
    \includegraphics[width=1\linewidth]{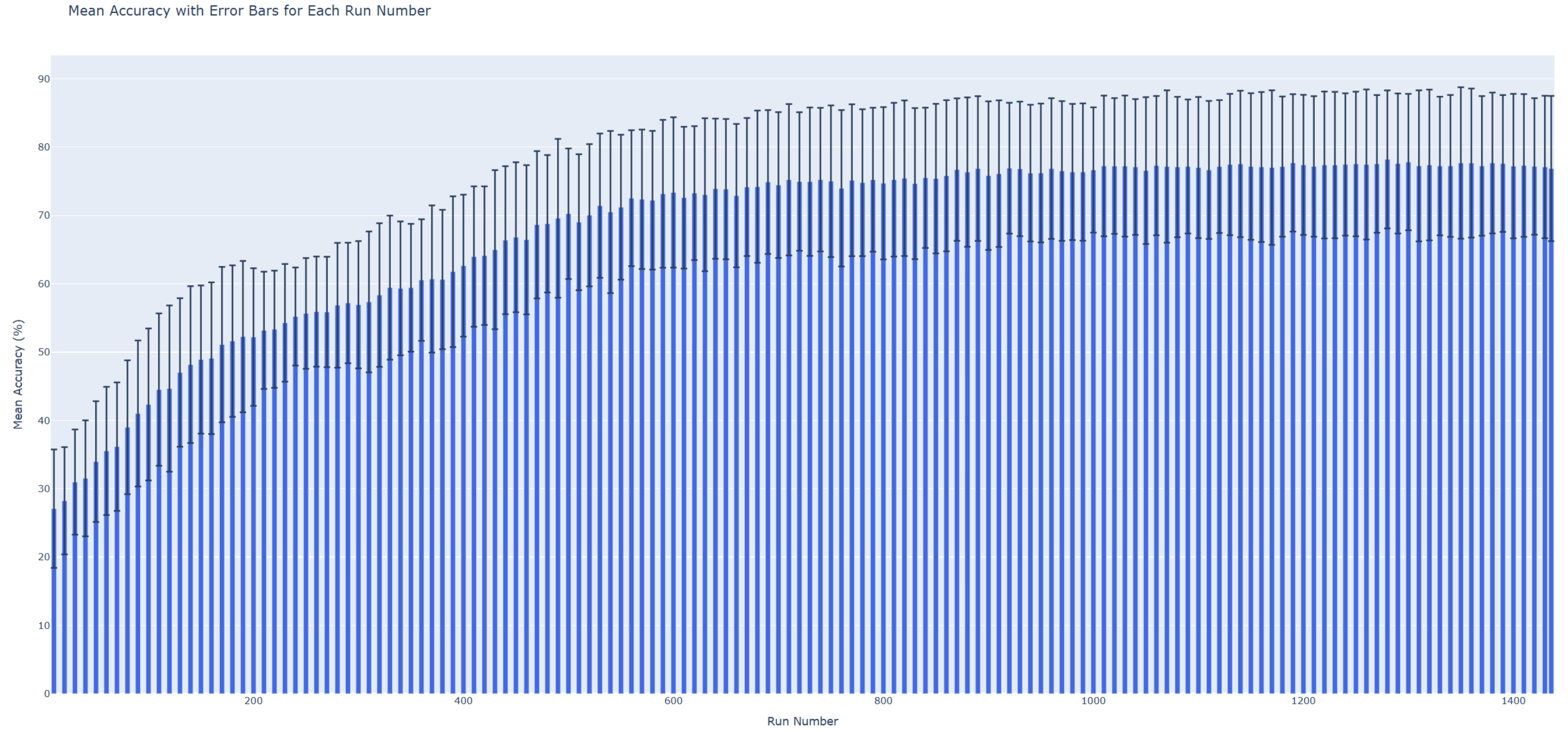}
    \caption{Bar chart depicting the average accuracy and standard deviation for the leave-one-out test in Experiment 1. After time step 520, the mean accuracy remains relatively stable, consistently exceeding 70\%. The standard deviation is around $\sigma = 10\%$ for most time steps, reflecting moderate variability in performance.}
    \label{fig:leave_Task1_barchart}
\end{figure*}

Figure~\ref{fig:leave_Task1_barchart} presents the mean accuracy and standard deviation of the test data across different window sizes. In Section~\ref{sec:task1_analysis}, we recommended the model with a window size of [0,380], achieving an accuracy of 71.09\%. However, under the leave-one-out test, the accuracy drops to $\mu = 60.62\% \, , \, \sigma = 10.19\%$. Notably, the mean accuracy remains relatively steady after time step 520, consistently staying above 70\%. Across most time steps, the standard deviation is approximately $\sigma = 10\%$, indicating moderate variability in performance.

\begin{figure*}[htb]
    \centering
    \includegraphics[width=1\linewidth]{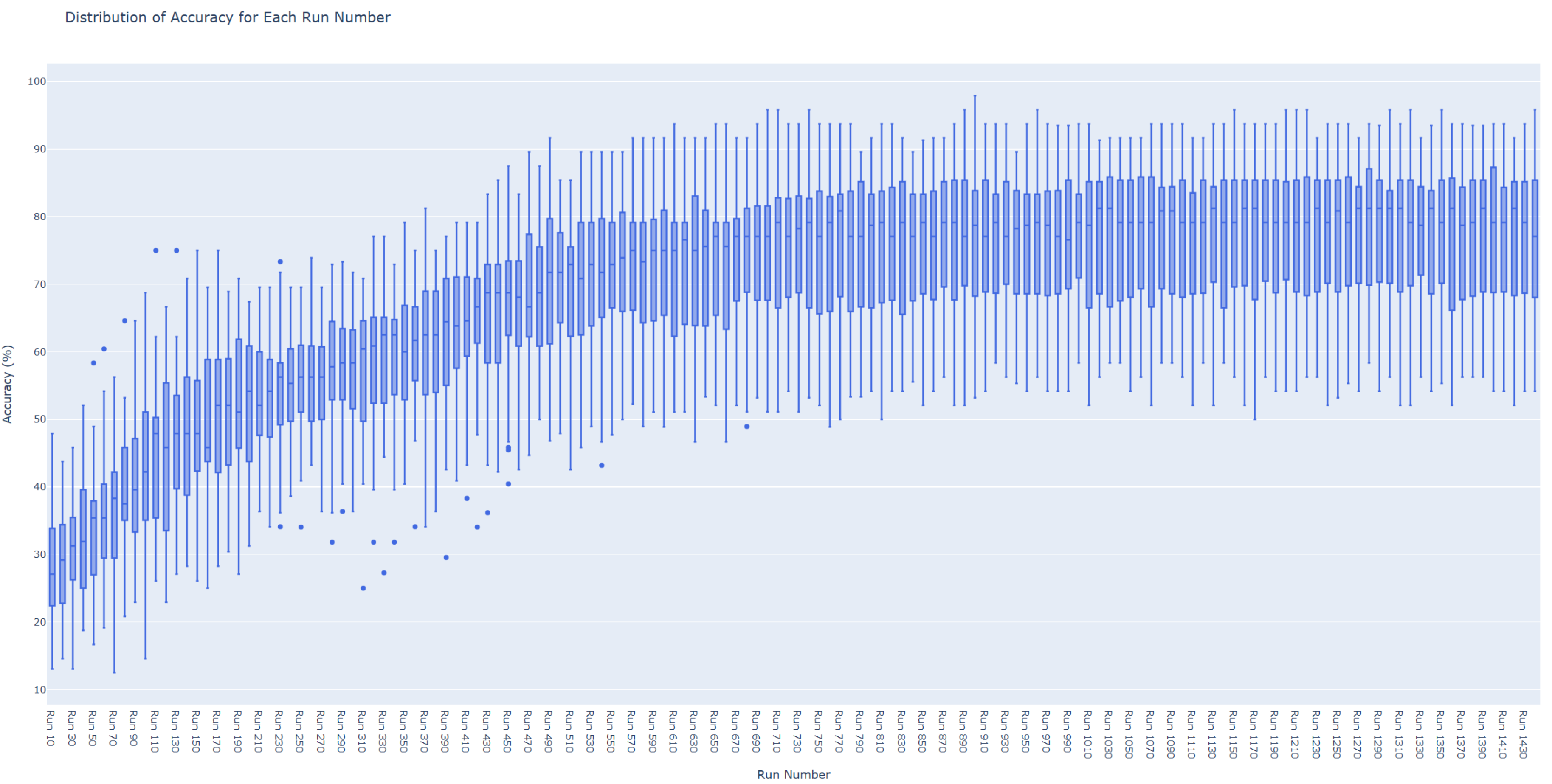}
    \caption{Box plot illustrating the accuracy distribution of each model in the leave-one-out test for Experiment 1. The proposed model with a window size of 380 achieves a median accuracy of $62.5\%$. After time step 490, the median accuracy consistently exceeds 70\%, eventually peaking at 81\% before settling at 77\% for the final model.}
    \label{fig:leave_Task1_boxplot}
\end{figure*}

To examine the distribution of each model's accuracy over time during the leave-one-out test, we plotted a boxplot of test accuracy across all models (see Figure~\ref{fig:leave_Task1_boxplot}). The results indicate that the suggested model with a window size of [0,380] has a median accuracy of $62.5\%$. Beyond time step 490, the median accuracy consistently remains above 70\%, eventually increasing to 81\% and achieving a median accuracy of 77\% for the final model.

\begin{figure*}[htb]
    \centering
    \includegraphics[width=1\linewidth]{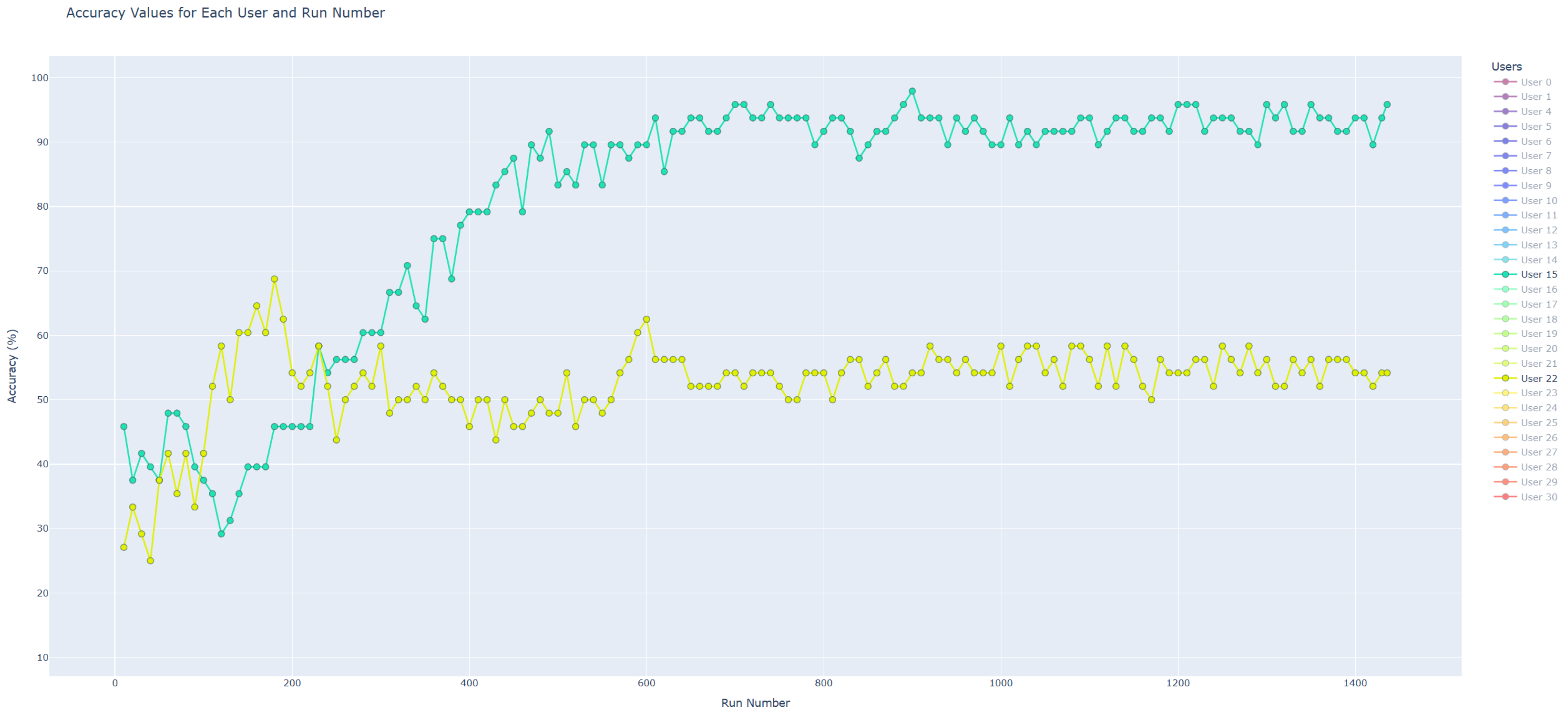}
    \caption{Effect of omitting individual users on model performance in the leave-one-out test for Experiment 1. The plot shows that excluding \textit{user22} significantly reduces model performance while removing \textit{user15} has minimal impact.}
    \label{fig:leave_Task1_colored_users_15_22}
\end{figure*}

To identify which users' exclusion from the training data significantly affects model performance and which have minimal impact, we plotted the accuracy of all models across all time steps, color-coded by the ID of the omitted user. The results reveal that removing \textit{user22} leads to a substantial drop in model performance, whereas omitting \textit{user15} has a negligible effect. These findings are illustrated in Figure~\ref{fig:leave_Task1_colored_users_15_22}. Hence, we conclude that \textit{user22} was (at least to some extent) an outlier, as it was hard to predict the objects that \textit{user22} was going to grasp from all the other participants.

\section{Discussion and Conclusion}
This paper explores the early prediction of multivariate time series data in a reach-to-grasp task, utilizing hand kinematic information. We present an interactive and explainable visual analysis tool, \textit{eXplainable Multivariate Time series Classification (XMTC)}, which enables users to analyze the multivariate time series, input features, prediction models, and distinguishing features. 

The classification of multivariate time series was performed using an interval-based ensemble classifier model DrCIF, adapted to our case study. The classification results were generally pleasing.
To assess generalizability, a leave-one-out test was performed for Experiment 1, showing a slight performance drop across models. The mean accuracy stabilizes after time step 520, consistently exceeding 70\%. Results highlight that removing \textit{user22} significantly impacts model performance, whereas omitting \textit{user15} has negligible effects.

The XMTC tool was developed in Python using Dash to create interactive web-based visualizations and facilitate user interaction. The tool provides valuable insights for identifying confusing classes over time, evaluating the model's accuracy on the test dataset, and examining the probability predictions for each class of a selected time series within the test dataset. Furthermore, it enables users to observe the global behavior of the model and identify distinguishing input features over time.

The developed XAI methods are global, post-hoc, and model-agnostic. Hence, they would also apply to any other classsification model. Thus, we could easily replace DrCIF with any other classification model for multi-variate time series.

The methods implemented in the XMTC tool have been applied to case studies from HCI looking into R2G sensor data and respectively defined aperture features. However, no assumptions are made that make the tool only applicable to such data. In future work, we would like to evaluate the performance and adaptability of XMTC across diverse domains, such as healthcare, finance, or human activity recognition, to demonstrate its generalization capability. Expanding its application to these fields could unlock new opportunities for early prediction and decision support in a wide range of real-world scenarios. Respective user studies across different domains may also provide valuable insight.

We showed in the case studies to R2G data that our tool facilitates the selection of an appropriate model based on the desired trade-off between accuracy and earliness, allowing for reliable prediction of the intended object before the grasping action is initiated. Such capabilities can significantly mitigate the effects of point-to-point latency in interactive environments, enhancing decision-making and user interaction.
In discussion with the domain expert, the XMTC was found suitable for the desired analysis and finding the desired model at the desired time step.
%
%Future work
Future research could focus on advancing action prediction by accurately predicting an action before it is performed. This capability could be particularly valuable in time-critical applications such as robotics, healthcare, and assistive technologies.

%Conducting user studies with participants who have expertise in visualization and machine learning could provide valuable insights into the usability and practicality of XMTC. These studies would help refine the tool’s interface and functionality based on user feedback, ensuring it accommodates the needs of a diverse audience, including both experts and non-specialists.

%Additionally, integrating advanced explainable AI (XAI) techniques, such as SHAP (SHapley Additive exPlanations) or LIME (Local Interpretable Model-Agnostic Explanations), could further enhance the interpretability of model predictions. These techniques can provide more granular insights into the contributions of individual features over time, improving user understanding of the underlying decision-making process.

\section*{Acknowledgments}
This project is funded by the Deutsche Forschungsgemeinschaft (DFG, German Research Foundation) – Project 436291335 and is part of Priority Program SPP2199 Scalable Interaction Paradigms for Pervasive Computing Environments.

% bibtex
\bibliographystyle{IEEEtran}  
\bibliography{refs}  
\end{document}